\documentclass[sigconf,balance]{acmart}
\AtBeginDocument{%
  \providecommand\BibTeX{{%
    \normalfont B\kern-0.5em{\scshape i\kern-0.25em b}\kern-0.8em\TeX}}}

\copyrightyear{2021} 
\acmYear{2021} 
\setcopyright{acmcopyright}\acmConference[MMAsia '21]{ACM Multimedia Asia}{December 1--3, 2021}{Gold Coast, Australia}
\acmBooktitle{ACM Multimedia Asia (MMAsia '21), December 1--3, 2021, Gold Coast, Australia}
\acmPrice{15.00}
\acmDOI{10.1145/3469877.3490600}
\acmISBN{978-1-4503-8607-4/21/12}

\usepackage{caption}
\usepackage{subcaption}

\usepackage{titlesec}
\usepackage{amsmath}
\usepackage{adjustbox}
\usepackage{hyperref}
\begin{document}

\title{Conditional Extreme Value Theory for Open Set Video Domain Adaptation}


\author{Zhuoxiao Chen, Yadan Luo, Mahsa Baktashmotlagh}
\affiliation{%
  \institution{The University of Queensland}
  \city{}
  \country{}}
\email{{zhuoxiao.chen, y.luo, m.baktashmotlagh}@uq.edu.au}


\begin{abstract}
With the advent of media streaming, video action recognition has become progressively important for various applications, yet at the high expense of requiring large-scale data labelling. To overcome the problem of expensive data labelling, domain adaptation techniques have been proposed, which transfer knowledge from fully labelled data (i.e., source domain) to unlabelled data (i.e., target domain). The majority of video domain adaptation algorithms are proposed for closed-set scenarios in which all the classes are shared among the domains. In this work, we propose an open-set video domain adaptation approach to mitigate the domain discrepancy between the source and target data, allowing the target data to contain additional classes that do not belong to the source domain. Different from previous works, which only focus on improving accuracy for shared classes, we aim to jointly enhance the alignment of the shared classes and recognition of unknown samples. Towards this goal, class-conditional extreme value theory is applied to enhance the unknown recognition. Specifically, the entropy values of target samples are modelled as generalised extreme value distributions, which allows separating unknown samples lying in the tail of the distribution. To alleviate the negative transfer issue, weights computed by the distance from the sample entropy to the threshold are leveraged in adversarial learning in the sense that confident source and target samples are aligned, and unconfident samples are pushed away. The proposed method has been thoroughly evaluated on both small-scale and large-scale cross-domain video datasets and achieved the state-of-the-art performance. 
\end{abstract}




\maketitle


\section{Introduction}
With the emergence of copious streaming media data, dynamically recognising and comprehending human actions and occurrence in online videos have become progressively essential, particularly for tasks like video content recommendation \cite{DBLP:conf/mm/WeiCYZZN19, DBLP:conf/mm/WeiWN0HC19}, surveillance \cite{DBLP:conf/mm/OuyangSZYS18}, and video retrieval \cite{DBLP:conf/mm/KwakHKH18}. Although supervised learning techniques \cite{DBLP:conf/nips/SimonyanZ14, DBLP:conf/iccv/TranBFTP15, wang2016temporal, DBLP:conf/cvpr/LiuWHDK17, DBLP:conf/iccv/RahmaniB17, DBLP:conf/cvpr/YanWLQ19} are beneficial for the tasks above, they lead to high expenses of labelling massive amounts of training data. The economical solution could be utilising a learner trained on existing labelled datasets to directly infer the labels of target datasets, yet there is often a domain shift between two datasets. Caused by the varying lighting conditions, camera angles and backgrounds, the domain shift triggers the performance drops of the learner. For example, synthetic video clips cropped from action-adventure games could be plentifully labelled and exploited, but inevitably has a huge domain shift from real-world videos such as action movie clips or sports video recordings. To address the issue of domain shift, unsupervised domain adaptation (UDA) techniques are introduced to align distributions between existing labelled data (source domain) and unlabelled data (target domain). To this end, existing UDA approaches either minimise the distribution distance across the domains \cite{DBLP:journals/jmlr/GrettonBRSS12, DBLP:conf/cvpr/YanDLWXZ17, DBLP:conf/iccv/BaktashmotlaghHLS13, DBLP:conf/ijcai/ErfaniBMNLBR16, DBLP:series/acvpr/BaktashmotlaghHS17} or learn the domain-invariant representations \cite{DBLP:conf/iccv/ChenKAYCZ19, DBLP:journals/pr/RahmanFBS20, DBLP:journals/corr/abs-2108-11726}.

 In the same vein, the video-based UDA methods aim to align the features at different levels such as frame, video, or temporal relation \cite{DBLP:conf/iccv/ChenKAYCZ19, DBLP:conf/bmvc/JamalNDV18, DBLP:conf/mm/LuoHW0B20}. However, existing video-based UDA methodologies fail to address an open-set scenario when target samples come from unknown classes that are not seen during training, and can cause negative transfer across domains. Thus, the ability to recognise the unknown classes and reject them from the domain alignment pipeline is essential to the open-set unsupervised domain adaptation (OUDA) task. Moreover, the existing OUDA frameworks are mainly evaluated on still image recognition datasets which are not effective enough to identify unknown samples when applied on video recognition benchmark datasets~\cite{DBLP:conf/icml/0002CZG19, DBLP:conf/icml/LuoWHB20, DBLP:conf/iclr/BaktashmotlaghF19, DBLP:conf/iccv/BustoG17, DBLP:conf/eccv/SaitoYUH18, DBLP:journals/corr/abs-1907-08375, DBLP:conf/icml/FangLLL021}.  


To overcome the above-mentioned limitations, we propose to intensify unknown recognition in open-set video domain adaptation. Our proposed framework consists of three modules. The first module is the Class-conditional Extreme Value Theory (CEVT) module that fits the entropy values of target samples to a set of generalised extreme value (GEV) distributions, where unknown samples can be efficiently identified as they lie on the tail of the distribution \cite{DBLP:journals/corr/abs-1811-08581, kotz2000extreme, DBLP:journals/pami/ZhangP17, DBLP:conf/cvpr/OzaP19}. Samples are fitted into the multiple class-conditional GEVs depending on the model's confidence in predicting those samples. For example, videos predicted as "pull-up" and "golf" are fitted into different GEVs. Then, we adaptively set a collection of thresholds for each GEV to split known and unknown samples. These fitted class-conditional GEVs with thresholds are employed in the other two modules. The second module is the class-conditional weighted domain adversarial learning pipeline to achieve the distribution alignment among shared classes and separate unknown classes. The weight of each sample is calculated by distance from entropy value to the threshold, which denotes the likelihood of belonging to the shared class or unknown class. At the inference stage, we have the third module of open-set recognition to classify samples with higher entropy than the threshold as the unknown class. This module in conjunction with class-conditional GEVs, is more robust to correctly classify hard classes than the typical approach of setting a global threshold. For example, in most of the existing OUDA approaches, the classifier predicts difficult "push-up" samples with the highest probability as "push-up" and with a lower probability as "pull-up". Subsequently, the entropy values for those samples are high, resulting in all the "push-up" samples getting rejected by the global threshold. However, in our framework, we fit all samples predicted as "push-up" to a GEV first, and then, we can efficiently separate "push-up" samples and unknown samples. This framework is particularly effective for complex video sets that the model encounters more challenging training and inference due to the complex spatio-temporal composition of video features. In general, our contributions are summarised as follows:
\begin{itemize}
  \item We propose a new Class-conditional Extreme Value Theory (CEVT) based framework for unsupervised video open-set domain adaptation that concentrates on domain-invariant representation learning via weighted domain adversarial learning.
  \item We investigate a new research direction of open-set video DA and introduce the CEVT model to solve the problem.
  \item Our proposed framework based on class-conditional extreme value theory is effective on both open-set recognition and adversarial weight generation, and it accurately recognises the unknown samples.
  \item We conducted extensive experiments to demonstrate the effectiveness of the proposed method on both small and large scale cross-domain video datasets and showed that the CEVT based framework achieves state-of-the-art performance. We released the source code of our proposed approach for reference: \textcolor{magenta}{ \url{https://github.com/zhuoxiao-chen/CEVT}}.
  
\end{itemize}

\begin{figure*}[ht]

    \centering
    \includegraphics[width=\textwidth]{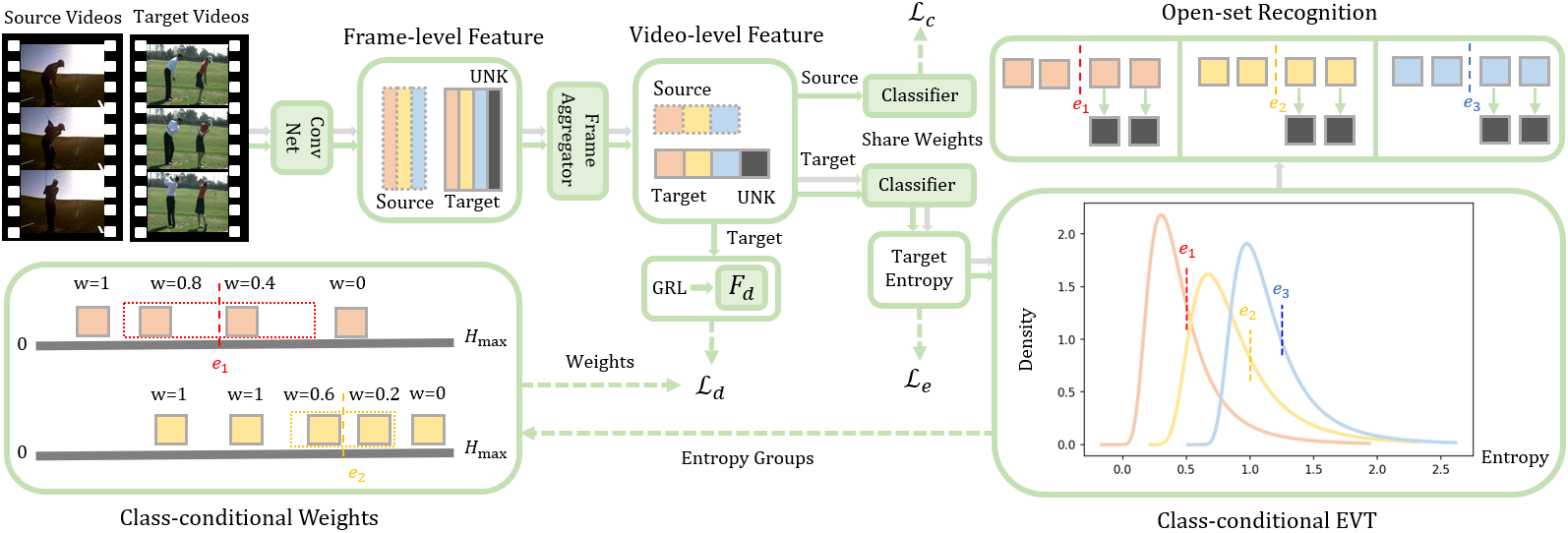}
    \caption{The overall  architecture of the proposed CEVT model. Different colours indicate different classes. Class-conditional EVT in the bottom right of the figure is the probability density function (PDF) of the Generalised Extreme Value (GEV) distribution for three different entropy groups. }
    \label{fig:model}
\end{figure*}

\section{RELATED WORK}
  
\subsection{Video Action Recognition}
Video action recognition is becoming increasingly important in the field of computer vision with many real-world applications, such as video surveillance \cite{DBLP:conf/mm/OuyangSZYS18}, video captioning and environment monitoring \cite{DBLP:conf/nips/DuanHGW0H18, DBLP:conf/iccv/KrishnaHRFN17, DBLP:conf/mm/WangWHWT18, DBLP:conf/mm/YangHW17}. To classify actions according to individual video frames or local motion vectors, a typical process employs a two-stream convolutional neural network \cite{DBLP:conf/cvpr/KarpathyTSLSF14, DBLP:conf/nips/SimonyanZ14}. Some works utilise attention \cite{DBLP:conf/cvpr/LongGM0LW18, DBLP:conf/cvpr/MaKMKAG18}, 3D convolutions \cite{DBLP:conf/iccv/TranBFTP15}, recurrent neural networks \cite{DBLP:journals/pami/DonahueHRVGSD17}, and temporal relation modules \cite{DBLP:conf/eccv/ZhouAOT18} to better extract long-term temporal features. Another branch of work, including 3D human skeleton recognition \cite{DBLP:conf/cvpr/LiuWHDK17, DBLP:conf/iccv/RahmaniB17}, complex object interactions \cite{DBLP:conf/cvpr/MaKMKAG18} and pose representations \cite{DBLP:conf/cvpr/YanWLQ19}, supplements the extracted RGB and optical flow features to alleviate the view dependency and noise caused by various lighting conditions. However, the above work necessitates costly annotations and could hardly be extended to an unseen situation, which significantly impedes the practical feasibility.
  
\subsection{Domain Adaptation}
To overcome such a limitation, Unsupervised Domain Adaptation (UDA) attempts to transfer knowledge from a labelled source domain to an unlabelled target domain. To tackle the domain shift referred to as the discrepancy of two domains, there are mainly two types of approaches. One is the discrepancy-based method that aims to minimise the distribution distance between two domains \cite{DBLP:journals/jmlr/GrettonBRSS12, DBLP:conf/cvpr/YanDLWXZ17, DBLP:conf/cvpr/BaktashmotlaghHLS14, DBLP:journals/jmlr/BaktashmotlaghH16, DBLP:conf/wacv/RahmanFBS19, DBLP:books/sp/20/RahmanFBS20}. The other is the adversary-based method that learns the domain-invariant representation \cite{DBLP:journals/jmlr/GaninUAGLLML16, DBLP:conf/mm/WangLHB20, DBLP:journals/corr/abs-2011-13549}. In addition, adversarial generative and self-supervision-based methods are also investigated by researchers \cite{9238468}. Recently, existing work has extended the UDA for harder video-based datasets. AMLS \cite{DBLP:conf/bmvc/JamalNDV18} applies pre-extracted C3D \cite{DBLP:conf/iccv/TranBFTP15} features to a Grassmann manifold derived from PCA and utilises adaptive kernels and adversarial learning to perform UDA. TA$^3$N \cite{DBLP:conf/iccv/ChenKAYCZ19} attempts to simultaneously align and learn temporal dynamics with entropy-based attention. Using the topology property of the bipartite graph network, ABG \cite{DBLP:conf/mm/LuoHW0B20} explicitly models source-target interactions to learn a domain-agnostic video classifier. Nevertheless, all methods mentioned above assume that the source domain and target domain share the same label set, which is not realistic in real-world scenarios. To address such issue, OUDA that assumes the target domain contains unknown classes, has made efforts at both theoretical and experimental level \cite{DBLP:conf/icml/0002CZG19, DBLP:conf/icml/LuoWHB20, DBLP:conf/iclr/BaktashmotlaghF19, DBLP:conf/iccv/BustoG17, DBLP:conf/eccv/BucciLT20, DBLP:conf/eccv/SaitoYUH18}. DMD \cite{DBLP:conf/icassp/WangSW0HC21} attempts to perform OUDA for videos, but fails to evaluate open recognition with the appropriate metric. Despite significant progress in a broader set of video classification and OUDA, domain adaptation has received little attention for knowledge transfer across videos under the open-set setting.

\section{METHODOLOGY}
In this section, we first give a formal definition of the Open-set Unsupervised Video Domain Adaptation (OUVDA), and then go through the details of our proposed CEVT framework, illustrated in Figure~\ref{fig:model}.

\subsection{Problem Formulation}
We are given a labelled source video set $\mathcal{S}=\{(X_i^s,y_i)\}_{i=1}^{N_s} \sim \mathbb{P}^s$ and an unlabelled target video collection $\mathcal{T}=\{(X_j^t)\}_{j=1}^{N_t} \sim \mathbb{Q}^t$, where $N_s$ and $N_t$ are the number of videos in each domain, respectively. Video samples in the source domain $\mathcal{S}$ and the target domain $\mathcal{T}$ are drawn from different probability distributions, \textit{i.e.}, $\mathbb{P}^s \neq \mathbb{Q}^t$. The two domains share $C$ common classes as the known classes. There is the additional class in the target domain $\mathcal{T}$ not shared with the source domain $\mathcal{S}$, which is regarded as the $C+1$, \textit{i.e.}, the unknown class. Each source video $X_i^s$ or target video $X_j^t$ is composed of $K$ frames, \textit{i.e.}, $X_i^s = {\{x_{ik}^s\}}_{k=1}^K$ and $X_j^t = {\{x_{jk}^t\}}_{k=1}^K$, where $x_{ik}^s, x_{jk}^t \in \mathbb{R}^D$ represent the $D$ dimensional feature vector of k-th frame in i-th source video and j-th target video, respectively. The primary goal of our method is to learn a classifier: $F_{\mathcal{T}}(\cdot;\theta_c)$ for predicting the $C$ labels of unlabelled videos in target domain and a collection of functions $F_{GEV}(\cdot;\mu, \sigma, \xi )$ to recognise samples from the unknown class $C+1$ in target domain.

\subsection{Source Classification}
We first feed both source videos and target videos into ResNet \cite{DBLP:journals/corr/HeZRS15} to obtain the frame-level features, \textit{i.e.}, $\{{\{x_{ik}^s\}}_{k=1}^K\}_{i=1}^{N_s}$ and $\{{\{x_{jk}^t\}}_{k=1}^K\}_{j=1}^{N_t}$ for source and target domain, respectively. Then the frame-level features are transformed into video-level features, \textit{i.e.}, $\{V_i^s\}_{i=1}^{N_s}$ and $\{V_j^t\}_{j=1}^{N_t}$ by frame aggregation techniques, with $V_i^s, V_i^t \in \mathbb{R}^D$. Without loss of generality, we utilise the mean Average Pooling (AvgPool), which is to produce a unified video representation by temporal averaging of the frame features. Thus, each source video-level feature $V_i^s$ and target video-level feature $V_j^t$ are defined as below:
\begin{align}
V_i^s = \frac{1}{K} \sum_{k=1}^{K} x_{ik}^s, V_j^t = \frac{1}{K} \sum_{k=1}^{K} x_{jk}^t.
\end{align}
Next, the aggregated features with labels $\{(V_i^s, y_i)\}_{i=1}^{N_s}$ from the source domain are fed into the source classifier $F_\mathcal{S}(\cdot;\theta_c)$, which is trained to minimise the cross entropy loss, 
\begin{align} \label{eq:cross_entropy_loss}
\mathcal{L}_c = -\mathbb{E}_{(V,y)\sim\mathbb{P}^s}  [\text{log}F_\mathcal{S}(V)].
\end{align}
The parameters of the source classifier $F_\mathcal{S}(\cdot;\theta_c)$ are shared with the target classifier $F_\mathcal{T}(\cdot;\theta_c)$, which is used to predict $C$ classes for target samples. Figure \ref{fig:model} shows the entire source classification process by a green array that starts from the videos on the left to the classification loss $\mathcal{L}_c$.

\subsection{Entropy-based Weights for Domain Adversarial Learning}
To align the distributions of the video-level source and target features, we propose a novel class-conditional EVT to generate conditional weights for domain adversarial learning by fitting the entropy values of target samples. The weighted domain adversarial learning can effectively align the known samples from both domains and separate the unknown samples from the target domain simultaneously by assigning instance-level weight to each sample. The samples which come by high probability from known classes are assigned a large weight. Conversely, samples that are likely to be unknown are given a small weight. Finally, all the features are multiplied by their weights and fed into the standard domain adversarial learning module $F_d(\cdot;\theta_d)$ with the Gradient Reversal Layer (GRL), as shown in Figure \ref{fig:model}.

\noindent\textbf{Class-conditional Extreme Value Theory}. To obtain the weight for each sample, we feed target samples $\{(V_j^t)\}_{j=1}^{N_t}$ into target classifier $F_\mathcal{T}(\cdot;\theta_c)$ to get the predictions for shared $C$ classes. Then, the entropy value of each target sample can be computed from its prediction,
\begin{align}
H(V^t) = -\sum_{c=1}^{C}
F_\mathcal{T}(V^t)_c\text{log}F_\mathcal{T}(V^t)_c,
\end{align}

\noindent where $F_\mathcal{T}(\cdot;\theta_c)_c$ denotes the prediction of class $c$ by the classifier. Then, the entropy values of target samples are partitioned into $C$ entropy groups, \textit{i.e.}, $G=\{G_i\}_{i=1}^{C}$. Target samples predicted to be from $i$-th class are allocated to the group $G_i$. The set of class-conditional entropy group is formulated as:
\begin{align}
G=\{H(V^t) :\text{argmax}(F_\mathcal{T}(V^t))=i \}_{i=1}^{C}.
\end{align}
\noindent Next, each group is fitted into a GEV distribution to obtain a set of CDFs of GEV, \textit{i.e.}, $\{F_{GEV}(\cdot \mid \text{class}=i)\}_{i=1}^{C}$, where $F_{GEV}(\cdot \mid \text{class}=i)$ indicates the CDF function fitted by entropy values in $i$-th group $G_i$. The CDF of GEV is calculated as,
\begin{align}
F_{GEV}(x;\mu, \sigma, \xi)=\text{exp}(-(1+\xi(\frac{x-\mu}{ \sigma}))^{-\frac{1}{\xi}}),
\end{align}
where $\mu$, $\sigma$, and $\xi$ are three parameters of GEV, determined by fitting data. Utilising the class-conditional EVT can complement the lack of class information for entropy. 

\noindent \textbf{Class-conditional Weights.} After fitting GEVs using entropy values of each entropy group, we set a global threshold $\delta$ for all the CDFs in $\{F_{GEV}(\cdot \mid \text{class}=i)\}_{i=1}^{C}$. Then, a set of class-conditional entropy threshold is computed for each group, denoted as,
\begin{align}
\{ e_i : F_{GEV}(e_i\mid \text{class}=i) = \delta \}_{i=1}^{C}.
\end{align}
Given the target sample, $V^t$ is classified in $i$-th class, if $H(V^t)$ is much greater or smaller than $e_i$, meaning this sample is very likely to be known or unknown. Then, we assign the weight as 1 or 0 to $V^t$. If the $H(V^t)$ is close to $e_i$, meaning the classifier is unsure about $H(V^t)$, we assign it a weight which linearly depends on distance from its entropy value to the corresponding class conditional entropy threshold $e_i$. The interval for linear variation of weight is named as mixture entropy interval, where most knowns or unknowns are mixed within this interval. The class-conditional weight can be formulated as,
\begin{align}
& W(V^t \mid \text{class}=i) = 0.5 + \frac{e_i-H(V^t)}{
\text{min}(e_i,(H_{max}(C) - e_i))
}, \\
& W(V^t \mid \text{class}=i) = \text{min}(1, \text{max}(0, W(V^t))),
\end{align}
where $H_{max}(C)$ is the entropy value of the evenly spaced vector with norm $|C|$, with each element of the vector being $1/C$. Two black line segments shown in the left bottom of Figure \ref{fig:model} shows the total entropy interval $[0, H_{max}]$, on which all the entropy values lie. The mixture entropy interval is $[e_i-\frac{\text{min}(e_i,(H_{max}(C) - e_i)}{2},e_i+\frac{\text{min}(e_i,(H_{max}(C) - e_i)}{2}]$, demonstrated by the dashed rectangles. The length of the entropy mixture interval varies depending on the distance from $e_i$ to either 0 or $H_{max}$, shown by red and yellow dashed rectangles of different length. The entropy threshold $e_i$ that is close to either 0 or $H_{max}$ indicates if the $i$-th class is easy or hard, and the distribution of class group $G_i$ has small variance. Thus, we need a small mixture for those dense group of entropy values to smoothly assign the weights. 

  

\noindent \textbf{Weighted Domain Adversarial Learning.} After obtaining weights for all the target samples by the class-conditional EVT technique, we train the domain classifier on the target video features multiplied with instance-level weights. The weighted domain classification loss is calculated by, 
\begin{align} \label{eq:adv_loss}
\begin{split}
\mathcal{L}_d & = \mathbb{E}_{V\sim\mathbb{P}^s}  [\text{log}F_d(V)] \\
 & + \mathbb{E}_{V\sim\mathbb{Q}^t}  [W(V\mid \text{class}=\text{argmax}(F_\mathcal{T}(V))\times\text{log}(1-F_d(V))].
 \end{split}
\end{align}
Gradually, with the proposed conditional EVT and weighted domain adversarial learning modules, known samples of both domains are aligned, and unknown samples of the target domain get separated from known samples.

\subsection{Entropy Maximisation}
To further separate the unknown samples, we utilise entropy maximum loss to progressively increase the entropy values of the overall target samples, defined as, 
\begin{align} \label{eq:entropy_loss}
\mathcal{L}_e & = -\mathbb{E}_{V\sim\mathbb{Q}^t}  [-\sum_{c=1}^{C}
F_\mathcal{T}(V)_c\text{log}F_\mathcal{T}(V)_c].
\end{align}
With the weighted domain adversarial learning, known target samples become similar to source samples. The entropy values of source samples gradually decreases because the source classifier is fully trained to optimise the cross-entropy loss $\mathcal{L}_c$. Thus, in the target domain, the entropy values of known samples decrease as well when optimising the two losses of $\mathcal{L}_c$ and  $\mathcal{L}_d$, while the entropy values of unknown samples increase when optimising $\mathcal{L}_e$ loss. Eventually, the unknown samples are optimally separated from known samples.

\begin{table}[!ht]
\caption{Performance comparisons on the UCF→HMDB.}
  \label{tab:UCF_to_HMDB}
  \resizebox{\linewidth}{!}{
  \begin{tabular}{lcccc|c}
    \toprule
    \textbf{Methods} & \textbf{ALL} & \textbf{OS} & \textbf{OS*} & \textbf{UNK} & \textbf{HOS}  \\
    \midrule
    
    DANN \cite{DBLP:journals/jmlr/GaninUAGLLML16} + OSVM \cite{DBLP:conf/eccv/JainSB14}  & 66.11 & 53.41 & 48.33 & 83.89 & 61.33\\
    
    JAN \cite{DBLP:conf/icml/LongZ0J17} + OSVM \cite{DBLP:conf/eccv/JainSB14}  & 61.11 & 51.59 & 47.78 & 74.44 & 58.20\\
    
    AdaBN \cite{DBLP:journals/pr/LiWSHL18} + OSVM \cite{DBLP:conf/eccv/JainSB14} & 60.65 & 61.35 & \textbf{61.67} & 59.44 & 60.54\\
    
    MCD \cite{DBLP:conf/cvpr/SaitoWUH18} + OSVM \cite{DBLP:conf/eccv/JainSB14}  & 66.67 & 60.32 & 57.78 & 75.56 & 65.48\\

    TA$^2$N \cite{DBLP:conf/iccv/ChenKAYCZ19} + OSVM \cite{DBLP:conf/eccv/JainSB14} & 65.28 & 58.73 & 56.11 & 74.44 & 63.99\\
    
    TA$^3$N \cite{DBLP:conf/iccv/ChenKAYCZ19} + OSVM \cite{DBLP:conf/eccv/JainSB14} & 62.25 & 55.95 & 53.33 & 71.67 & 61.16 \\
    
    OSBP \cite{DBLP:conf/eccv/SaitoYUH18} + AvgPool & 67.19 & 55.64 & 50.83 & 84.47 & 63.47\\\hline
    \textbf{Ours}  & \textbf{75.28} & \textbf{61.59} & 56.11 & \textbf{94.44} & \textbf{70.40} \\
  \bottomrule
\end{tabular}
}
\vspace{-10pt}
\end{table}

\begin{table}[!ht]
\caption{Performance comparisons on the HMDB→UCF.}
  \label{tab:HMDB_to_UCF}
  \resizebox{\linewidth}{!}{
  \begin{tabular}{l cccc|c}
    \toprule
    \textbf{Methods}  & \textbf{ALL} & \textbf{OS} & \textbf{OS*} & \textbf{UNK} & \textbf{HOS}  \\
    \midrule
    
    DANN \cite{DBLP:journals/jmlr/GaninUAGLLML16} + OSVM \cite{DBLP:conf/eccv/JainSB14} &  64.62 & 64.61 & 62.94 & 74.67 & 68.31 \\
    
    JAN \cite{DBLP:conf/icml/LongZ0J17} + OSVM \cite{DBLP:conf/eccv/JainSB14} &  61.47 & 64.47 & 62.91 & 73.80 & 67.92 \\
    
    AdaBN \cite{DBLP:journals/pr/LiWSHL18} + OSVM \cite{DBLP:conf/eccv/JainSB14}  & 62.87 & 60.86 & 58.78 & 73.36 & 65.27 \\
    
    MCD \cite{DBLP:conf/cvpr/SaitoWUH18} + OSVM \cite{DBLP:conf/eccv/JainSB14} & 66.73 & 64.96 & 63.48 & 73.80 & 68.25 \\
    
    TA$^2$N \cite{DBLP:conf/iccv/ChenKAYCZ19} + OSVM \cite{DBLP:conf/eccv/JainSB14}  & 63.40 & 63.88 & 61.35 & 79.04 & 69.08 \\
    
    TA$^3$N \cite{DBLP:conf/iccv/ChenKAYCZ19} + OSVM \cite{DBLP:conf/eccv/JainSB14} & 60.60 & 61.84 & 58.39 & 82.53 & 68.39 \\

    OSBP \cite{DBLP:conf/eccv/SaitoYUH18} + AvgPool &  64.84 & 59.61 & 55.26 & \textbf{85.71} & 67.19 \\\hline 
    
    \textbf{Ours} & \textbf{70.58} & \textbf{69.29} & \textbf{66.79} & 84.28 &\textbf{74.52} \\
  \bottomrule
\end{tabular}
}
\end{table}

\subsection{Optimisation}
The ultimate objective is to learn the optimal parameters for the CEVT model,
\begin{align}
\begin{split}
&(\theta_c^*,\theta_e^*)=\text{argmin}\: \mathcal{L}_c+\beta\mathcal{L}_e - \gamma\mathcal{L}_d,\\
&(\theta_d^*)=\text{argmin}\: \mathcal{L}_c+\beta\mathcal{L}_e+\gamma\mathcal{L}_d,
\end{split}
\end{align}
with $\beta$ and $\gamma$ the coefficients of the entropy maximisation loss and weighted adversarial loss, respectively.

 \subsection{Inference}
In this section, we explain the inference stage of the proposed CEVT after the model parameters are optimised. The inference process is denoted by grey arrays shown in Figure \ref{fig:model}. The target videos are fed into the convolution network, frame aggregator, and the target classifier $F_\mathcal{T}(\cdot;\theta_c^*)$. Then, the predictions are passed into the class-conditional EVT module for open-set recognition. The predicted class $y \in |C+1|$ of each input target videos is represented as,
\begin{align}
\begin{split}
&y=F_\mathcal{T}(V^t) \\
&y=C+1 \:\: \text{if} \; F_{GEV}(V_t\mid\text{class}=\text{argmax}(F_\mathcal{T}(V^t)))>\delta.
\end{split}
\end{align}


\section{EXPERIMENTS}
In this section, we empirically evaluate the performance of the proposed CEVT model on two datasets, \textbf{UCF-HMDB} and \textbf{UCF-Olympic} for unsupervised open-set domain adaptation.

\subsection{Datasets}

The \textbf{UCF-HMDB} is the intersected subset covering 12 highly relevant categories of two large-scale video action datasets, the UCF101 \cite{DBLP:journals/corr/abs-1212-0402} and HMDB51 \cite{DBLP:conf/iccv/KuehneJGPS11}, including \textit{Climb, Fencing, Golf, Kick Ball, pull-up, Punch, push-up, Ride Bike, Ride Horse, Shoot Ball, Shoot Bow and Walk}. The \textbf{UCF-Olympic} have six common categories from the UCF101 and Olympic Sports Dataset \cite{DBLP:conf/eccv/NieblesCF10}, which involves \textit{Basketball, Clearn and Jerk, Diving, Pole Vault, Tennis and Discus Throw}. These dataset partitioning strategies follow \cite{DBLP:conf/iccv/ChenKAYCZ19} to make a fair comparison. Likewise, we utilise the pre-extracted frame-level features by ResNet101 model pre-trained on ImageNet. In terms of known/unknown category splitting, we select the first half categories as known classes, and all the remaining categories are labelled as unknown, for both \textbf{UCF-HMDB} and \textbf{UCF-Olympic}.

\subsection{Evaluation Metrics}
To compare the performance of the proposed \textbf{CEVT} and the baseline methods, we adopt four widely used metrics \cite{DBLP:conf/eccv/BucciLT20}\cite{DBLP:conf/eccv/SaitoYUH18} for evaluating OUDA tasks. The accuracy (\textbf{ALL}) is the correctly predicted target samples over all target samples. \textbf{OS} is the average class accuracy over the classes. \textbf{OS*} is the average class accuracy over the known classes. \textbf{UNK} is the unknown class accuracy. \textbf{HOS} is the harmonic mean of \textbf{OS*} and \textbf{UNK} formulated as: 
$
2\times \frac
{\textbf{OS}^* \times \textbf{UNK}}
{\textbf{OS}^* + \textbf{UNK}}
$ 
. \textbf{HOS} is the most meaningful metric for evaluating OUDA tasks because it can best reflect the balance between \textbf{OS*} and \textbf{UNK}. 


\subsection{Baselines}
We compare our proposed \textbf{CEVT} with three types of state-of-the-art domain adaptation methods: the close-set method for images, the close-set method for videos and the open-set method for images. The close-set domain adaptation methods for images are extended to align the distributions of aggregated frames from source and target domains, which include Domain-Adversarial Neural Network (\textbf{DANN}) \cite{DBLP:journals/jmlr/GaninUAGLLML16}, Joint Adaptation Network (\textbf{JAN}) \cite{DBLP:conf/icml/LongZ0J17}, Adaptive Batch Normalisation (\textbf{AdaBN}) \cite{DBLP:journals/pr/LiWSHL18} and Maximum Classier Discrepancy (\textbf{MCD}) \cite{DBLP:conf/cvpr/SaitoWUH18}. In terms of the close-set approach for videos, Temporal Attentive Adversarial Adaptation Network (\textbf{TA$^2$N}) \cite{DBLP:conf/iccv/ChenKAYCZ19} and Temporal Adversarial Adaptation Network (\textbf{TA$^3$N}) \cite{DBLP:conf/iccv/ChenKAYCZ19} are adopted for comparison. We equip the above open-set methods with the OSVM \cite{DBLP:conf/eccv/JainSB14} for open recognition. As for the open-set method for images, OUDA by Backpropagation (\textbf{OSBP}) \cite{DBLP:conf/eccv/SaitoYUH18} extended by frame aggregator is made into comparison.

\subsection{Implementation Details}
All the baselines and our approach are implemented by PyTorch \cite{DBLP:conf/nips/PaszkeGMLBCKLGA19} on one server with two GeForce GTX 2080 Ti GPUs. We follow \cite{DBLP:conf/iccv/ChenKAYCZ19} to sample a specified number of frames with uniform spacing from each video for training and extract a 2048-D feature vector from each frame by the Resnet-101 pre-trained on ImageNet. To ensure a fair comparison, we fix $K$ to 16 for all methods using average pooling, and follow the optimisation strategy in \cite{DBLP:conf/iccv/ChenKAYCZ19} to utilise the stochastic gradient descent (SGD) as the optimiser, and learning-rate-decreasing techniques from \textbf{DANN} \cite{DBLP:journals/jmlr/GaninUAGLLML16}, with learning rate, momentum, and weight decay of 0.03, 0.9 and $1\times10^{-4}$, respectively. The scale of datasets determines the size of the source batch, 32 for UCF-Olympic and 128 for UCF-HMDB. The size of the target batch is computed by multiplying the source batch with the ratio between the source and target datasets. The loss coefficient $\beta$ is set as 1, 10, 0.19, 0.22, and  $\gamma$ is set as 0.9, 0.7, 1.83, 5, for UCF→HMDB, HMDB→UCF, UCF→Olympic and Olympic→UCF, respectively. The EVT threshold is set as 0.4, 0.45, 0.6 and 0.29 for the above tasks, respectively.

\begin{table}[t]
\caption{Performance comparisons on the Olympic→UCF.}
  \label{tab:O_to_U}
  \resizebox{\linewidth}{!}{
  \begin{tabular}{lcccc|c}
    \toprule
    \textbf{Methods} & \textbf{ALL} & \textbf{OS} & \textbf{OS*} & \textbf{UNK} & \textbf{HOS}  \\
    \midrule
    
    DANN \cite{DBLP:journals/jmlr/GaninUAGLLML16} + OSVM \cite{DBLP:conf/eccv/JainSB14}  & 83.33 & 84.93 & \textbf{86.38} & 80.60 & 83.39 \\
    
    JAN \cite{DBLP:conf/icml/LongZ0J17} + OSVM \cite{DBLP:conf/eccv/JainSB14} & 88.75 & 84.23 & 80.46 & \textbf{95.52} & 87.35 \\
    
    AdaBN \cite{DBLP:journals/pr/LiWSHL18} + OSVM \cite{DBLP:conf/eccv/JainSB14} & 84.17 & 80.09 & 76.93 & 89.55 & 82.76\\
    
    MCD \cite{DBLP:conf/cvpr/SaitoWUH18} + OSVM \cite{DBLP:conf/eccv/JainSB14}  & 83.75 & 84.65 & 85.50 & 82.09 & 83.76 \\

    TA$^2$N \cite{DBLP:conf/iccv/ChenKAYCZ19} + OSVM \cite{DBLP:conf/eccv/JainSB14} & 87.92 & 82.74 & 78.48 & \textbf{95.52} & 86.17 \\
    
    TA$^3$N \cite{DBLP:conf/iccv/ChenKAYCZ19} + OSVM \cite{DBLP:conf/eccv/JainSB14} & 85.83 & 85.64 & 85.58 & 85.82 & 85.70 \\
    
    OSBP \cite{DBLP:conf/eccv/SaitoYUH18} + AvgPool & 89.06 & 86.23 & 84.31 & 92.00 & 87.98 \\\hline
    \textbf{Ours}  & \textbf{89.17} & \textbf{87.54} & \textbf{86.38} & 91.04 &  \textbf{88.65} \\

  \bottomrule
\end{tabular}
}
\end{table}

\begin{table}[t]
\caption{Performance comparisons on the UCF→Olympic.}
  \label{tab:U_to_O}
  \resizebox{\linewidth}{!}{
  \begin{tabular}{lcccc|c}
    \toprule
    \textbf{Methods} & \textbf{ALL} & \textbf{OS} & \textbf{OS*} & \textbf{UNK} & \textbf{HOS}  \\
    \midrule
    
    DANN \cite{DBLP:journals/jmlr/GaninUAGLLML16} + OSVM \cite{DBLP:conf/eccv/JainSB14}  & 94.44 & 95.33 & 96.67 & 91.30 & 93.91 \\
    
    JAN \cite{DBLP:conf/icml/LongZ0J17} + OSVM \cite{DBLP:conf/eccv/JainSB14}  & 94.44 & 96.74 & \textbf{100.00} & 86.96 & 93.02 \\
    
    AdaBN \cite{DBLP:journals/pr/LiWSHL18} + OSVM \cite{DBLP:conf/eccv/JainSB14} & 87.04 & 83.86 & 78.48 & \textbf{100.00} & 87.95 \\
    
    MCD \cite{DBLP:conf/cvpr/SaitoWUH18} + OSVM \cite{DBLP:conf/eccv/JainSB14}  & 87.04 & 86.74 & 86.67 & 86.96 & 86.81 \\

    TA$^2$N \cite{DBLP:conf/iccv/ChenKAYCZ19} + OSVM \cite{DBLP:conf/eccv/JainSB14} & 96.30 & \textbf{97.83} & \textbf{100.00} & 91.30 & 95.45 \\
    
    TA$^3$N \cite{DBLP:conf/iccv/ChenKAYCZ19} + OSVM \cite{DBLP:conf/eccv/JainSB14} & 88.89 & 88.74 & 87.88 & 91.03 & 89.56 \\
    
    OSBP \cite{DBLP:conf/eccv/SaitoYUH18} + AvgPool & 96.88 & 95.83 & 94.44 & \textbf{100.00} & 97.14 \\\hline
    \textbf{Ours}  & \textbf{98.15} & 97.73 & 96.97 & \textbf{100.00} & \textbf{98.46} \\

  \bottomrule
\end{tabular}
}\vspace{-10pt}
\end{table}

\subsection{Comparisons with State-of-The-Art}
We clearly report the performance of the proposed \textbf{CEVT} and baseline methods on \textbf{UCF-HMDB} and \textbf{UCF-Olympic} as shown in Table \ref{tab:UCF_to_HMDB}, Table \ref{tab:HMDB_to_UCF}, Table \ref{tab:O_to_U} and Table \ref{tab:U_to_O}. The proposed \textbf{CEVT} model outperforms all the compared state-of-the-art domain adaptation approaches, improving the \textbf{HOS} by 4.92\%, 5.44\%, 1.32\% and 0.67\% on the adaptation task UCF→HMDB, HMDB→UCF, UCF→Olympic and Olympic→UCF, respectively. It is worth noting that the proposed model achieves significant performance boosts for the larger-scale datasets of \textbf{UCF-Olympic}. Also, note that the outstanding performance gain of the proposed framework for the most challenging transfer task, \textit{i.e.}, UCF→HMDB, illustrates the better adaptation ability of our approach. Some methods achieve 100\% on \textbf{OS*} or \textbf{UNK} for the UCF→Olympic task because this task is relatively easier than other tasks, and the validation set has limited samples. Also, there is usually a trade-off between these two metrics. For example, \textbf{JAN} achieves 100\% on \textbf{OS*} but gets the lowest score (86.95\%) on \textbf{UNK}. Conversely, \textbf{AdaBn} achieves 100\% on \textbf{UNK} but has the worst performance (78.48\%) on \textbf{OS*}. The proposed \textbf{CEVT} is superior to all baselines, as it achieves remarkably high \textbf{OS*} and \textbf{UNK} simultaneously.

\begin{figure*}[ht]

    \centering
    \begin{subfigure}[t]{0.22\textwidth}
        \centering
        \includegraphics[height=1.3in]{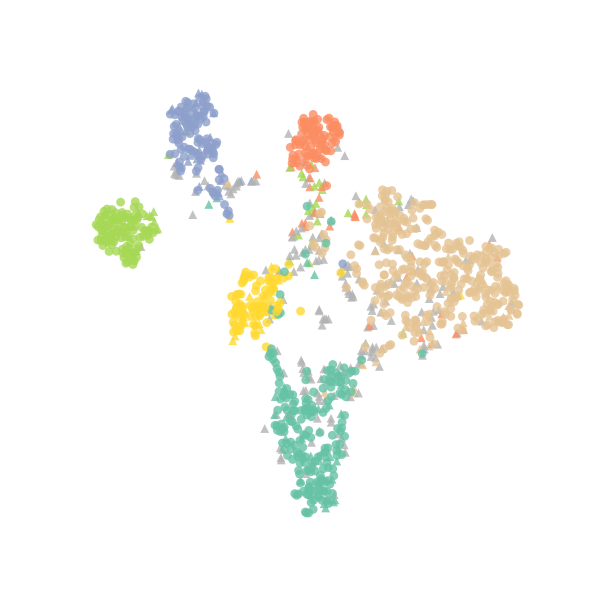}
        \caption{DANN + OSVM}
    \end{subfigure}%
    ~
    \begin{subfigure}[t]{0.22\textwidth}
        \centering
        \includegraphics[height=1.3in]{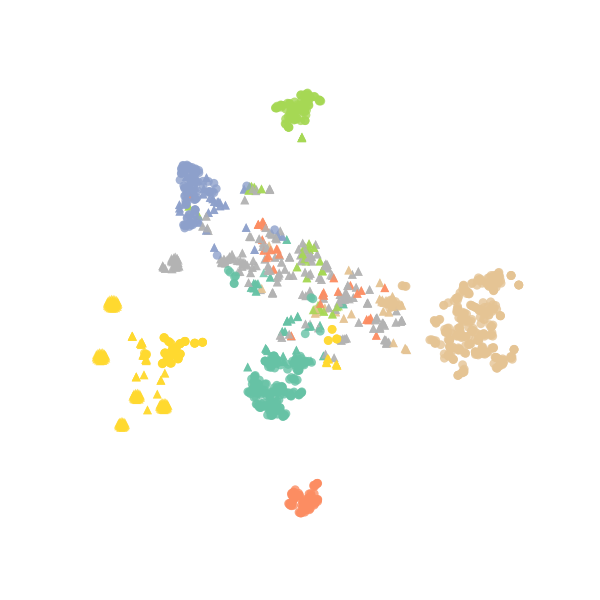}
        \caption{OSBP + AvgPool}
    \end{subfigure}%
    ~
        \begin{subfigure}[t]{0.22\textwidth}
        \centering
        \includegraphics[height=1.3in]{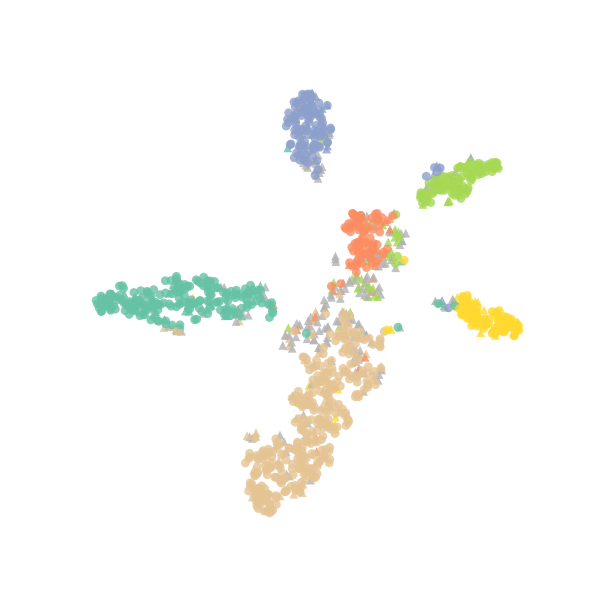}
        \caption{TA$^3$N + OSVM}
    \end{subfigure}%
    ~
    \begin{subfigure}[t]{0.22\textwidth}
        \centering
        \includegraphics[height=1.3in]{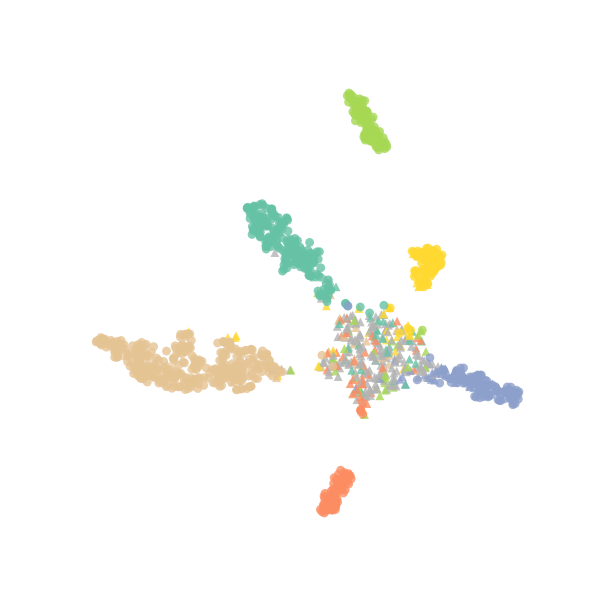}
        \caption{Ours}
    \end{subfigure}

    \caption{The t-SNE visualisation of the learned source and target video representations on the UCF→HMDB task.}
    \label{fig:tsne}
\end{figure*}

\begin{table}[ht]
\caption{The ablation performance (HOS\%) of the proposed CEVT model on the UCF-HMDB dataset. "w" indicates with and "w/o" indicates without.}
  \label{tab:ablation_study}
  \begin{tabular}{lcl}
    \toprule
    \textbf{Methods} & \textbf{UCF→HMDB} & \textbf{HMDB→UCF}\\
    \midrule
    CEVT w/o $\mathcal{L}_{e}$ \& $\mathcal{L}_{d}$ & 62.09 & 71.42\\
    CEVT w/o $\mathcal{L}_{e}$  & 64.04 & 72.01\\
    CEVT w/o $\mathcal{L}_{d}$  & 67.41 & 72.88 \\
    CEVT w unweighted $\mathcal{L}_{d}$ & 68.03 & 72.56 \\
    CEVT  & \textbf{70.40} & \textbf{74.52} \\
  \bottomrule
\end{tabular}
\end{table}

\subsection{Ablation Study}
We use the UCF-HMDB dataset to investigate the performance of the proposed modules of the model CEVT. Table \ref{tab:ablation_study} summarises the experimental results among methods with different removed functions. Removing both weighted domain adversarial learning and entropy maximisation learning, the performance of \textbf{CEVT w/o $\mathcal{L}_{e}$} drop (8.31\%) significantly when adapting from UCF to HMDB compared with the complete model. Although there is only source classification and EVT method for unknown recognition left, the dropped performance (62.09\% and 71.42\%) is still better than most of the baseline methods equipped with OSVM, proving the robustness of EVT in terms of unknown recognition tasks. The \textbf{CEVT w/o $\mathcal{L}_{e}$} refers to the variant without the entropy maximisation loss, which causes HOS drop (6.36\% and 1.09\% ) in both adaptation directions. Removing the weighted adversarial learning, referred to as \textbf{CEVT w/o $\mathcal{L}_{d}$}, result in a performance decrease slightly in either direction. The  \textbf{CEVT w unweighted $\mathcal{L}_{d}$} indicates that the weights of all instances are set as 1 for $\mathcal{L}_{d}$, which performs worse than weighted $\mathcal{L}_{d}$. The transfer task from UCF to HMDB is more challenging than the opposite direction. Thus, the proposed modules are more capable of handling challenging adaptation tasks. 

\subsection{Parameter Sensitivity}
\begin{figure}[ht]

  \centering
  \includegraphics[width=\linewidth]{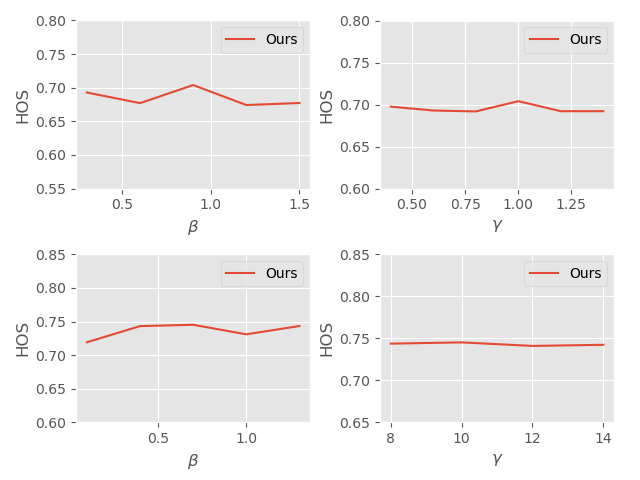}
  \caption{Performance (HOS) comparisons of the proposed CEVT with respect to the varying loss coefficients on the UCF→HMDB (shown in the upper row) and HMDB→UCF (shown in the bottom row) adaptation tasks.}
  \label{fig:praram_sensitivity}
\end{figure}

To explore the sensitivity of the loss coefficients of the proposed CEVT, we run the experiments on the dataset \textbf{UCF-HMDB} with changing values of $\beta$ and $\gamma$, which are used to adjust the weighted adversarial loss and the entropy maximisation loss, respectively. Even though both $\beta$ and $\gamma$ vary in a wide interval, as plotted in Figure \ref{fig:praram_sensitivity}, the average HOS of the proposed CEVT is very steady for both \textbf{UCF→HMDB} and \textbf{HMDB→UCF} tasks. No matter how the coefficients $\beta$ varies, the fluctuation range of HOS does not exceed $3\%$. As for the $\gamma$, the fluctuation range is less than $1.5\%$. This demonstrates the resistance of our methodology to varying loss coefficients.
  \vspace{0.5cm}

\vspace{-20pt}
\subsection{Visualisation}
To intuitively show how our model closes the domain shift between source and target domains as well as effectively recognises the unknown samples, we apply the t-SNE \cite{JMLR:v9:vandermaaten08a} to visualise the extracted features from baseline models, \textbf{DANN}, \textbf{OSPB}, \textbf{TA$^3$N} and our proposed \textbf{CEVT} on the UCF→HMDB task as shown in Figure \ref{fig:tsne}. Different colours denote different classes, and unknown samples are grey. The source videos are represented by circles, while triangles represent the target videos. Compared to the baseline methods, it is noticeable that the features extracted by CEVT produce tighter clusters, and unknown samples are better clustered to the centre.

\section{CONCLUSION}
In this work, we propose a CEVT framework to tackle the problem of open-set unsupervised video domain adaptation. Unlike previous works, we intensify the open recognition to jointly improve the accuracy of the both known and unknown classes. Experiments demonstrate that the proposed algorithm outperforms state-of-the-art methods on both large and small video datasets. Future work includes testing CEVT for still images, and equipping the proposed CEVT with other frame aggregators for videos.

\newpage

\bibliographystyle{ACM-Reference-Format}
\bibliography{main}


\begin{thebibliography}{60}


\ifx \showCODEN    \undefined \def \showCODEN     #1{\unskip}     \fi
\ifx \showDOI      \undefined \def \showDOI       #1{#1}\fi
\ifx \showISBNx    \undefined \def \showISBNx     #1{\unskip}     \fi
\ifx \showISBNxiii \undefined \def \showISBNxiii  #1{\unskip}     \fi
\ifx \showISSN     \undefined \def \showISSN      #1{\unskip}     \fi
\ifx \showLCCN     \undefined \def \showLCCN      #1{\unskip}     \fi
\ifx \shownote     \undefined \def \shownote      #1{#1}          \fi
\ifx \showarticletitle \undefined \def \showarticletitle #1{#1}   \fi
\ifx \showURL      \undefined \def \showURL       {\relax}        \fi
\providecommand\bibfield[2]{#2}
\providecommand\bibinfo[2]{#2}
\providecommand\natexlab[1]{#1}
\providecommand\showeprint[2][]{arXiv:#2}

\bibitem[\protect\citeauthoryear{Baktashmotlagh, Faraki, Drummond, and
  Salzmann}{Baktashmotlagh et~al\mbox{.}}{2019}]%
        {DBLP:conf/iclr/BaktashmotlaghF19}
\bibfield{author}{\bibinfo{person}{Mahsa Baktashmotlagh},
  \bibinfo{person}{Masoud Faraki}, \bibinfo{person}{Tom Drummond}, {and}
  \bibinfo{person}{Mathieu Salzmann}.} \bibinfo{year}{2019}\natexlab{}.
\newblock \showarticletitle{Learning Factorized Representations for Open-Set
  Domain Adaptation}. In \bibinfo{booktitle}{\emph{Proc. International
  Conference on Learning Representations, {ICLR} 2019}}.
\newblock


\bibitem[\protect\citeauthoryear{Baktashmotlagh, Harandi, Lovell, and
  Salzmann}{Baktashmotlagh et~al\mbox{.}}{2013}]%
        {DBLP:conf/iccv/BaktashmotlaghHLS13}
\bibfield{author}{\bibinfo{person}{Mahsa Baktashmotlagh},
  \bibinfo{person}{Mehrtash~Tafazzoli Harandi}, \bibinfo{person}{Brian~C.
  Lovell}, {and} \bibinfo{person}{Mathieu Salzmann}.}
  \bibinfo{year}{2013}\natexlab{}.
\newblock \showarticletitle{Unsupervised Domain Adaptation by Domain Invariant
  Projection}. In \bibinfo{booktitle}{\emph{Proc. International Conference on
  Computer Vision, {ICCV} 2013}}. \bibinfo{publisher}{{IEEE} Computer Society},
  \bibinfo{pages}{769--776}.
\newblock


\bibitem[\protect\citeauthoryear{Baktashmotlagh, Harandi, Lovell, and
  Salzmann}{Baktashmotlagh et~al\mbox{.}}{2014}]%
        {DBLP:conf/cvpr/BaktashmotlaghHLS14}
\bibfield{author}{\bibinfo{person}{Mahsa Baktashmotlagh},
  \bibinfo{person}{Mehrtash~Tafazzoli Harandi}, \bibinfo{person}{Brian~C.
  Lovell}, {and} \bibinfo{person}{Mathieu Salzmann}.}
  \bibinfo{year}{2014}\natexlab{}.
\newblock \showarticletitle{Domain Adaptation on the Statistical Manifold}. In
  \bibinfo{booktitle}{\emph{Proc. Conference on Computer Vision and Pattern
  Recognition, {CVPR} 2014}}. \bibinfo{pages}{2481--2488}.
\newblock


\bibitem[\protect\citeauthoryear{Baktashmotlagh, Harandi, and
  Salzmann}{Baktashmotlagh et~al\mbox{.}}{2016}]%
        {DBLP:journals/jmlr/BaktashmotlaghH16}
\bibfield{author}{\bibinfo{person}{Mahsa Baktashmotlagh},
  \bibinfo{person}{Mehrtash~Tafazzoli Harandi}, {and} \bibinfo{person}{Mathieu
  Salzmann}.} \bibinfo{year}{2016}\natexlab{}.
\newblock \showarticletitle{Distribution-Matching Embedding for Visual Domain
  Adaptation}.
\newblock \bibinfo{journal}{\emph{Journal of Machine Learning Research}}
  \bibinfo{volume}{17} (\bibinfo{year}{2016}), \bibinfo{pages}{108:1--108:30}.
\newblock


\bibitem[\protect\citeauthoryear{Baktashmotlagh, Harandi, and
  Salzmann}{Baktashmotlagh et~al\mbox{.}}{2017}]%
        {DBLP:series/acvpr/BaktashmotlaghHS17}
\bibfield{author}{\bibinfo{person}{Mahsa Baktashmotlagh},
  \bibinfo{person}{Mehrtash~Tafazzoli Harandi}, {and} \bibinfo{person}{Mathieu
  Salzmann}.} \bibinfo{year}{2017}\natexlab{}.
\newblock \showarticletitle{Learning Domain Invariant Embeddings by Matching
  Distributions}.
\newblock In \bibinfo{booktitle}{\emph{Domain Adaptation in Computer Vision
  Applications}}. \bibinfo{pages}{95--114}.
\newblock


\bibitem[\protect\citeauthoryear{Bucci, Loghmani, and Tommasi}{Bucci
  et~al\mbox{.}}{2020}]%
        {DBLP:conf/eccv/BucciLT20}
\bibfield{author}{\bibinfo{person}{Silvia Bucci},
  \bibinfo{person}{Mohammad~Reza Loghmani}, {and} \bibinfo{person}{Tatiana
  Tommasi}.} \bibinfo{year}{2020}\natexlab{}.
\newblock \showarticletitle{On the Effectiveness of Image Rotation for Open Set
  Domain Adaptation}. In \bibinfo{booktitle}{\emph{Proc. European Conference on
  Computer Vision, {ECCV} 2020}}. \bibinfo{pages}{422--438}.
\newblock


\bibitem[\protect\citeauthoryear{Busto and Gall}{Busto and Gall}{2017}]%
        {DBLP:conf/iccv/BustoG17}
\bibfield{author}{\bibinfo{person}{Pau~Panareda Busto} {and}
  \bibinfo{person}{Juergen Gall}.} \bibinfo{year}{2017}\natexlab{}.
\newblock \showarticletitle{Open Set Domain Adaptation}. In
  \bibinfo{booktitle}{\emph{Proc. International Conference on Computer Vision,
  {ICCV} 2017}}. \bibinfo{pages}{754--763}.
\newblock


\bibitem[\protect\citeauthoryear{Chen, Kira, Alregib, Yoo, Chen, and
  Zheng}{Chen et~al\mbox{.}}{2019}]%
        {DBLP:conf/iccv/ChenKAYCZ19}
\bibfield{author}{\bibinfo{person}{Min{-}Hung Chen}, \bibinfo{person}{Zsolt
  Kira}, \bibinfo{person}{Ghassan Alregib}, \bibinfo{person}{Jaekwon Yoo},
  \bibinfo{person}{Ruxin Chen}, {and} \bibinfo{person}{Jian Zheng}.}
  \bibinfo{year}{2019}\natexlab{}.
\newblock \showarticletitle{Temporal Attentive Alignment for Large-Scale Video
  Domain Adaptation}. In \bibinfo{booktitle}{\emph{Proc. International
  Conference on Computer Vision, {ICCV} 2019}}. \bibinfo{pages}{6320--6329}.
\newblock


\bibitem[\protect\citeauthoryear{Donahue, Hendricks, Rohrbach, Venugopalan,
  Guadarrama, Saenko, and Darrell}{Donahue et~al\mbox{.}}{2017}]%
        {DBLP:journals/pami/DonahueHRVGSD17}
\bibfield{author}{\bibinfo{person}{Jeff Donahue}, \bibinfo{person}{Lisa~Anne
  Hendricks}, \bibinfo{person}{Marcus Rohrbach}, \bibinfo{person}{Subhashini
  Venugopalan}, \bibinfo{person}{Sergio Guadarrama}, \bibinfo{person}{Kate
  Saenko}, {and} \bibinfo{person}{Trevor Darrell}.}
  \bibinfo{year}{2017}\natexlab{}.
\newblock \showarticletitle{Long-Term Recurrent Convolutional Networks for
  Visual Recognition and Description}.
\newblock \bibinfo{journal}{\emph{IEEE Transactions on Pattern Analysis and
  Machine Intelligence}} \bibinfo{volume}{39}, \bibinfo{number}{4}
  (\bibinfo{year}{2017}), \bibinfo{pages}{677--691}.
\newblock


\bibitem[\protect\citeauthoryear{Duan, Huang, Gan, Wang, Zhu, and Huang}{Duan
  et~al\mbox{.}}{2018}]%
        {DBLP:conf/nips/DuanHGW0H18}
\bibfield{author}{\bibinfo{person}{Xuguang Duan}, \bibinfo{person}{Wen{-}bing
  Huang}, \bibinfo{person}{Chuang Gan}, \bibinfo{person}{Jingdong Wang},
  \bibinfo{person}{Wenwu Zhu}, {and} \bibinfo{person}{Junzhou Huang}.}
  \bibinfo{year}{2018}\natexlab{}.
\newblock \showarticletitle{Weakly Supervised Dense Event Captioning in
  Videos}. In \bibinfo{booktitle}{\emph{Proc. Conference on Neural Information
  Processing Systems, NeurIPS 2018}}. \bibinfo{pages}{3063--3073}.
\newblock


\bibitem[\protect\citeauthoryear{Erfani, Baktashmotlagh, Moshtaghi, Nguyen,
  Leckie, Bailey, and Ramamohanarao}{Erfani et~al\mbox{.}}{2016}]%
        {DBLP:conf/ijcai/ErfaniBMNLBR16}
\bibfield{author}{\bibinfo{person}{Sarah~M. Erfani}, \bibinfo{person}{Mahsa
  Baktashmotlagh}, \bibinfo{person}{Masud Moshtaghi}, \bibinfo{person}{Vinh
  Nguyen}, \bibinfo{person}{Christopher Leckie}, \bibinfo{person}{James
  Bailey}, {and} \bibinfo{person}{Kotagiri Ramamohanarao}.}
  \bibinfo{year}{2016}\natexlab{}.
\newblock \showarticletitle{Robust Domain Generalisation by Enforcing
  Distribution Invariance}. In \bibinfo{booktitle}{\emph{Proc. International
  Joint Conference on Artificial Intelligence, {IJCAI} 2016}}.
  \bibinfo{pages}{1455--1461}.
\newblock


\bibitem[\protect\citeauthoryear{Fang, Lu, Liu, Liu, and Zhang}{Fang
  et~al\mbox{.}}{2021}]%
        {DBLP:conf/icml/FangLLL021}
\bibfield{author}{\bibinfo{person}{Zhen Fang}, \bibinfo{person}{Jie Lu},
  \bibinfo{person}{Anjin Liu}, \bibinfo{person}{Feng Liu}, {and}
  \bibinfo{person}{Guangquan Zhang}.} \bibinfo{year}{2021}\natexlab{}.
\newblock \showarticletitle{Learning Bounds for Open-Set Learning}. In
  \bibinfo{booktitle}{\emph{Proc. of the 37th International Conference on
  Machine Learning, {ICML} 2021}}. \bibinfo{pages}{3122--3132}.
\newblock


\bibitem[\protect\citeauthoryear{Fang, Lu, Liu, Xuan, and Zhang}{Fang
  et~al\mbox{.}}{2019}]%
        {DBLP:journals/corr/abs-1907-08375}
\bibfield{author}{\bibinfo{person}{Zhen Fang}, \bibinfo{person}{Jie Lu},
  \bibinfo{person}{Feng Liu}, \bibinfo{person}{Junyu Xuan}, {and}
  \bibinfo{person}{Guangquan Zhang}.} \bibinfo{year}{2019}\natexlab{}.
\newblock \showarticletitle{Open Set Domain Adaptation: Theoretical Bound and
  Algorithm}.
\newblock \bibinfo{journal}{\emph{IEEE Transactions on Neural Networks and
  Learning Systems}} (\bibinfo{year}{2019}).
\newblock


\bibitem[\protect\citeauthoryear{Ganin, Ustinova, Ajakan, Germain, Larochelle,
  Laviolette, Marchand, and Lempitsky}{Ganin et~al\mbox{.}}{2016}]%
        {DBLP:journals/jmlr/GaninUAGLLML16}
\bibfield{author}{\bibinfo{person}{Yaroslav Ganin}, \bibinfo{person}{Evgeniya
  Ustinova}, \bibinfo{person}{Hana Ajakan}, \bibinfo{person}{Pascal Germain},
  \bibinfo{person}{Hugo Larochelle}, \bibinfo{person}{Fran{\c{c}}ois
  Laviolette}, \bibinfo{person}{Mario Marchand}, {and}
  \bibinfo{person}{Victor~S. Lempitsky}.} \bibinfo{year}{2016}\natexlab{}.
\newblock \showarticletitle{Domain-Adversarial Training of Neural Networks}.
\newblock \bibinfo{journal}{\emph{Journal of Machine Learning Research}}
  \bibinfo{volume}{17} (\bibinfo{year}{2016}), \bibinfo{pages}{59:1--59:35}.
\newblock


\bibitem[\protect\citeauthoryear{Geng, Huang, and Chen}{Geng
  et~al\mbox{.}}{2018}]%
        {DBLP:journals/corr/abs-1811-08581}
\bibfield{author}{\bibinfo{person}{Chuanxing Geng},
  \bibinfo{person}{Sheng{-}Jun Huang}, {and} \bibinfo{person}{Songcan Chen}.}
  \bibinfo{year}{2018}\natexlab{}.
\newblock \showarticletitle{Recent Advances in Open Set Recognition: {A}
  Survey}.
\newblock \bibinfo{journal}{\emph{CoRR}}  \bibinfo{volume}{abs/1811.08581}
  (\bibinfo{year}{2018}).
\newblock


\bibitem[\protect\citeauthoryear{Gretton, Borgwardt, Rasch, Sch{\"{o}}lkopf,
  and Smola}{Gretton et~al\mbox{.}}{2012}]%
        {DBLP:journals/jmlr/GrettonBRSS12}
\bibfield{author}{\bibinfo{person}{Arthur Gretton}, \bibinfo{person}{Karsten~M.
  Borgwardt}, \bibinfo{person}{Malte~J. Rasch}, \bibinfo{person}{Bernhard
  Sch{\"{o}}lkopf}, {and} \bibinfo{person}{Alexander~J. Smola}.}
  \bibinfo{year}{2012}\natexlab{}.
\newblock \showarticletitle{A Kernel Two-Sample Test}.
\newblock \bibinfo{journal}{\emph{Journal of Machine Learning Research}}
  \bibinfo{volume}{13} (\bibinfo{year}{2012}), \bibinfo{pages}{723--773}.
\newblock


\bibitem[\protect\citeauthoryear{He, Zhang, Ren, and Sun}{He
  et~al\mbox{.}}{2015}]%
        {DBLP:journals/corr/HeZRS15}
\bibfield{author}{\bibinfo{person}{Kaiming He}, \bibinfo{person}{Xiangyu
  Zhang}, \bibinfo{person}{Shaoqing Ren}, {and} \bibinfo{person}{Jian Sun}.}
  \bibinfo{year}{2015}\natexlab{}.
\newblock \showarticletitle{Deep Residual Learning for Image Recognition}.
\newblock \bibinfo{journal}{\emph{CoRR}}  \bibinfo{volume}{abs/1512.03385}
  (\bibinfo{year}{2015}).
\newblock


\bibitem[\protect\citeauthoryear{Jain, Scheirer, and Boult}{Jain
  et~al\mbox{.}}{2014}]%
        {DBLP:conf/eccv/JainSB14}
\bibfield{author}{\bibinfo{person}{Lalit~P. Jain}, \bibinfo{person}{Walter~J.
  Scheirer}, {and} \bibinfo{person}{Terrance~E. Boult}.}
  \bibinfo{year}{2014}\natexlab{}.
\newblock \showarticletitle{Multi-class Open Set Recognition Using Probability
  of Inclusion}. In \bibinfo{booktitle}{\emph{Proc. European Conference on
  Computer Vision, {ECCV} 2014}}. \bibinfo{pages}{393--409}.
\newblock


\bibitem[\protect\citeauthoryear{Jamal, Namboodiri, Deodhare, and
  Venkatesh}{Jamal et~al\mbox{.}}{2018}]%
        {DBLP:conf/bmvc/JamalNDV18}
\bibfield{author}{\bibinfo{person}{Arshad Jamal}, \bibinfo{person}{Vinay~P.
  Namboodiri}, \bibinfo{person}{Dipti Deodhare}, {and} \bibinfo{person}{K.~S.
  Venkatesh}.} \bibinfo{year}{2018}\natexlab{}.
\newblock \showarticletitle{Deep Domain Adaptation in Action Space}. In
  \bibinfo{booktitle}{\emph{Proc. British Machine Vision Conference, {BMVC}
  2018}}. \bibinfo{pages}{264}.
\newblock


\bibitem[\protect\citeauthoryear{Karpathy, Toderici, Shetty, Leung, Sukthankar,
  and Li}{Karpathy et~al\mbox{.}}{2014}]%
        {DBLP:conf/cvpr/KarpathyTSLSF14}
\bibfield{author}{\bibinfo{person}{Andrej Karpathy}, \bibinfo{person}{George
  Toderici}, \bibinfo{person}{Sanketh Shetty}, \bibinfo{person}{Thomas Leung},
  \bibinfo{person}{Rahul Sukthankar}, {and} \bibinfo{person}{Fei{-}Fei Li}.}
  \bibinfo{year}{2014}\natexlab{}.
\newblock \showarticletitle{Large-Scale Video Classification with Convolutional
  Neural Networks}. In \bibinfo{booktitle}{\emph{Proc. Conference on Computer
  Vision and Pattern Recognition, {CVPR} 2014}}. \bibinfo{pages}{1725--1732}.
\newblock


\bibitem[\protect\citeauthoryear{Kotz and Nadarajah}{Kotz and
  Nadarajah}{2000}]%
        {kotz2000extreme}
\bibfield{author}{\bibinfo{person}{Samuel Kotz} {and} \bibinfo{person}{Saralees
  Nadarajah}.} \bibinfo{year}{2000}\natexlab{}.
\newblock \bibinfo{booktitle}{\emph{Extreme value distributions: theory and
  applications}}.
\newblock \bibinfo{publisher}{World Scientific}.
\newblock


\bibitem[\protect\citeauthoryear{Krishna, Hata, Ren, Fei{-}Fei, and
  Niebles}{Krishna et~al\mbox{.}}{2017}]%
        {DBLP:conf/iccv/KrishnaHRFN17}
\bibfield{author}{\bibinfo{person}{Ranjay Krishna}, \bibinfo{person}{Kenji
  Hata}, \bibinfo{person}{Frederic Ren}, \bibinfo{person}{Li Fei{-}Fei}, {and}
  \bibinfo{person}{Juan~Carlos Niebles}.} \bibinfo{year}{2017}\natexlab{}.
\newblock \showarticletitle{Dense-Captioning Events in Videos}. In
  \bibinfo{booktitle}{\emph{Proc. International Conference on Computer Vision,
  {ICCV} 2017}}. \bibinfo{pages}{706--715}.
\newblock


\bibitem[\protect\citeauthoryear{Kuehne, Jhuang, Garrote, Poggio, and
  Serre}{Kuehne et~al\mbox{.}}{2011}]%
        {DBLP:conf/iccv/KuehneJGPS11}
\bibfield{author}{\bibinfo{person}{Hildegard Kuehne}, \bibinfo{person}{Hueihan
  Jhuang}, \bibinfo{person}{Est{\'{\i}}baliz Garrote},
  \bibinfo{person}{Tomaso~A. Poggio}, {and} \bibinfo{person}{Thomas Serre}.}
  \bibinfo{year}{2011}\natexlab{}.
\newblock \showarticletitle{{HMDB:} {A} large video database for human motion
  recognition}. In \bibinfo{booktitle}{\emph{Proc. International Conference on
  Computer Vision, {ICCV} 2011}}. \bibinfo{pages}{2556--2563}.
\newblock


\bibitem[\protect\citeauthoryear{Kwak, Han, Kim, and Hahm}{Kwak
  et~al\mbox{.}}{2018}]%
        {DBLP:conf/mm/KwakHKH18}
\bibfield{author}{\bibinfo{person}{Chang{-}Uk Kwak}, \bibinfo{person}{Minho
  Han}, \bibinfo{person}{Sun{-}Joong Kim}, {and} \bibinfo{person}{Gyeong{-}June
  Hahm}.} \bibinfo{year}{2018}\natexlab{}.
\newblock \showarticletitle{Interactive Story Maker: Tagged Video Retrieval
  System for Video Re-creation Service}. In \bibinfo{booktitle}{\emph{Proc.
  International Conference on Multimedia, {MM} 2018}}.
  \bibinfo{pages}{1270--1271}.
\newblock


\bibitem[\protect\citeauthoryear{Li, Wang, Shi, Hou, and Liu}{Li
  et~al\mbox{.}}{2018}]%
        {DBLP:journals/pr/LiWSHL18}
\bibfield{author}{\bibinfo{person}{Yanghao Li}, \bibinfo{person}{Naiyan Wang},
  \bibinfo{person}{Jianping Shi}, \bibinfo{person}{Xiaodi Hou}, {and}
  \bibinfo{person}{Jiaying Liu}.} \bibinfo{year}{2018}\natexlab{}.
\newblock \showarticletitle{Adaptive Batch Normalization for practical domain
  adaptation}.
\newblock \bibinfo{journal}{\emph{Pattern Recognition}}  \bibinfo{volume}{80}
  (\bibinfo{year}{2018}), \bibinfo{pages}{109--117}.
\newblock


\bibitem[\protect\citeauthoryear{Liu, Wang, Hu, Duan, and Kot}{Liu
  et~al\mbox{.}}{2017}]%
        {DBLP:conf/cvpr/LiuWHDK17}
\bibfield{author}{\bibinfo{person}{Jun Liu}, \bibinfo{person}{Gang Wang},
  \bibinfo{person}{Ping Hu}, \bibinfo{person}{Ling{-}Yu Duan}, {and}
  \bibinfo{person}{Alex~C. Kot}.} \bibinfo{year}{2017}\natexlab{}.
\newblock \showarticletitle{Global Context-Aware Attention {LSTM} Networks for
  3D Action Recognition}. In \bibinfo{booktitle}{\emph{Proc. Conference on
  Computer Vision and Pattern Recognition, {CVPR} 2017}}.
  \bibinfo{pages}{3671--3680}.
\newblock


\bibitem[\protect\citeauthoryear{Long, Zhu, Wang, and Jordan}{Long
  et~al\mbox{.}}{2017}]%
        {DBLP:conf/icml/LongZ0J17}
\bibfield{author}{\bibinfo{person}{Mingsheng Long}, \bibinfo{person}{Han Zhu},
  \bibinfo{person}{Jianmin Wang}, {and} \bibinfo{person}{Michael~I. Jordan}.}
  \bibinfo{year}{2017}\natexlab{}.
\newblock \showarticletitle{Deep Transfer Learning with Joint Adaptation
  Networks}. In \bibinfo{booktitle}{\emph{Proc. International Conference on
  Machine Learning, {ICML} 2017}}. \bibinfo{pages}{2208--2217}.
\newblock


\bibitem[\protect\citeauthoryear{Long, Gan, de~Melo, Wu, Liu, and Wen}{Long
  et~al\mbox{.}}{2018}]%
        {DBLP:conf/cvpr/LongGM0LW18}
\bibfield{author}{\bibinfo{person}{Xiang Long}, \bibinfo{person}{Chuang Gan},
  \bibinfo{person}{Gerard de Melo}, \bibinfo{person}{Jiajun Wu},
  \bibinfo{person}{Xiao Liu}, {and} \bibinfo{person}{Shilei Wen}.}
  \bibinfo{year}{2018}\natexlab{}.
\newblock \showarticletitle{Attention Clusters: Purely Attention Based Local
  Feature Integration for Video Classification}. In
  \bibinfo{booktitle}{\emph{Proc. Conference on Computer Vision and Pattern
  Recognition, {CVPR} 2018}}. \bibinfo{pages}{7834--7843}.
\newblock


\bibitem[\protect\citeauthoryear{Luo, Huang, Wang, Zhang, and
  Baktashmotlagh}{Luo et~al\mbox{.}}{2020a}]%
        {DBLP:conf/mm/LuoHW0B20}
\bibfield{author}{\bibinfo{person}{Yadan Luo}, \bibinfo{person}{Zi Huang},
  \bibinfo{person}{Zijian Wang}, \bibinfo{person}{Zheng Zhang}, {and}
  \bibinfo{person}{Mahsa Baktashmotlagh}.} \bibinfo{year}{2020}\natexlab{a}.
\newblock \showarticletitle{Adversarial Bipartite Graph Learning for Video
  Domain Adaptation}. In \bibinfo{booktitle}{\emph{Proc. International
  Conference on Multimedia, MM 2020}}. \bibinfo{pages}{19--27}.
\newblock


\bibitem[\protect\citeauthoryear{Luo, Wang, Huang, and Baktashmotlagh}{Luo
  et~al\mbox{.}}{2020b}]%
        {DBLP:conf/icml/LuoWHB20}
\bibfield{author}{\bibinfo{person}{Yadan Luo}, \bibinfo{person}{Zijian Wang},
  \bibinfo{person}{Zi Huang}, {and} \bibinfo{person}{Mahsa Baktashmotlagh}.}
  \bibinfo{year}{2020}\natexlab{b}.
\newblock \showarticletitle{Progressive Graph Learning for Open-Set Domain
  Adaptation}. In \bibinfo{booktitle}{\emph{Proc. of the 37th International
  Conference on Machine Learning, {ICML} 2020}}. \bibinfo{pages}{6468--6478}.
\newblock


\bibitem[\protect\citeauthoryear{Ma, Kadav, Melvin, Kira, AlRegib, and Graf}{Ma
  et~al\mbox{.}}{2018}]%
        {DBLP:conf/cvpr/MaKMKAG18}
\bibfield{author}{\bibinfo{person}{Chih{-}Yao Ma}, \bibinfo{person}{Asim
  Kadav}, \bibinfo{person}{Iain Melvin}, \bibinfo{person}{Zsolt Kira},
  \bibinfo{person}{Ghassan AlRegib}, {and} \bibinfo{person}{Hans~Peter Graf}.}
  \bibinfo{year}{2018}\natexlab{}.
\newblock \showarticletitle{Attend and Interact: Higher-Order Object
  Interactions for Video Understanding}. In \bibinfo{booktitle}{\emph{Proc.
  Conference on Computer Vision and Pattern Recognition, {CVPR} 2018}}.
  \bibinfo{pages}{6790--6800}.
\newblock


\bibitem[\protect\citeauthoryear{Moghimifar, Haffari, and
  Baktashmotlagh}{Moghimifar et~al\mbox{.}}{2020}]%
        {DBLP:journals/corr/abs-2011-13549}
\bibfield{author}{\bibinfo{person}{Farhad Moghimifar},
  \bibinfo{person}{Gholamreza Haffari}, {and} \bibinfo{person}{Mahsa
  Baktashmotlagh}.} \bibinfo{year}{2020}\natexlab{}.
\newblock \showarticletitle{Domain Adaptative Causality Encoder}.
\newblock \bibinfo{journal}{\emph{CoRR}}  \bibinfo{volume}{abs/2011.13549}
  (\bibinfo{year}{2020}).
\newblock


\bibitem[\protect\citeauthoryear{Niebles, Chen, and Li}{Niebles
  et~al\mbox{.}}{2010}]%
        {DBLP:conf/eccv/NieblesCF10}
\bibfield{author}{\bibinfo{person}{Juan~Carlos Niebles},
  \bibinfo{person}{Chih{-}Wei Chen}, {and} \bibinfo{person}{Fei{-}Fei Li}.}
  \bibinfo{year}{2010}\natexlab{}.
\newblock \showarticletitle{Modeling Temporal Structure of Decomposable Motion
  Segments for Activity Classification}. In \bibinfo{booktitle}{\emph{Proc.
  European Conference on Computer Vision, {ECCV} 2010}}.
  \bibinfo{pages}{392--405}.
\newblock


\bibitem[\protect\citeauthoryear{Ouyang, Shao, Zhang, Yang, and Shen}{Ouyang
  et~al\mbox{.}}{2018}]%
        {DBLP:conf/mm/OuyangSZYS18}
\bibfield{author}{\bibinfo{person}{Deqiang Ouyang}, \bibinfo{person}{Jie Shao},
  \bibinfo{person}{Yonghui Zhang}, \bibinfo{person}{Yang Yang}, {and}
  \bibinfo{person}{Heng~Tao Shen}.} \bibinfo{year}{2018}\natexlab{}.
\newblock \showarticletitle{Video-based Person Re-identification via Self-Paced
  Learning and Deep Reinforcement Learning Framework}. In
  \bibinfo{booktitle}{\emph{Proc. International Conference on Multimedia, {MM}
  2018}}. \bibinfo{pages}{1562--1570}.
\newblock


\bibitem[\protect\citeauthoryear{Oza and Patel}{Oza and Patel}{2019}]%
        {DBLP:conf/cvpr/OzaP19}
\bibfield{author}{\bibinfo{person}{Poojan Oza} {and} \bibinfo{person}{Vishal~M.
  Patel}.} \bibinfo{year}{2019}\natexlab{}.
\newblock \showarticletitle{{C2AE:} Class Conditioned Auto-Encoder for Open-Set
  Recognition}. In \bibinfo{booktitle}{\emph{Proc. Conference on Computer
  Vision and Pattern Recognition, {CVPR} 2019}}. \bibinfo{pages}{2307--2316}.
\newblock


\bibitem[\protect\citeauthoryear{Paszke, Gross, Massa, Lerer, Bradbury, Chanan,
  Killeen, Lin, Gimelshein, Antiga, Desmaison, K{\"{o}}pf, Yang, DeVito,
  Raison, Tejani, Chilamkurthy, Steiner, Fang, Bai, and Chintala}{Paszke
  et~al\mbox{.}}{2019}]%
        {DBLP:conf/nips/PaszkeGMLBCKLGA19}
\bibfield{author}{\bibinfo{person}{Adam Paszke}, \bibinfo{person}{Sam Gross},
  \bibinfo{person}{Francisco Massa}, \bibinfo{person}{Adam Lerer},
  \bibinfo{person}{James Bradbury}, \bibinfo{person}{Gregory Chanan},
  \bibinfo{person}{Trevor Killeen}, \bibinfo{person}{Zeming Lin},
  \bibinfo{person}{Natalia Gimelshein}, \bibinfo{person}{Luca Antiga},
  \bibinfo{person}{Alban Desmaison}, \bibinfo{person}{Andreas K{\"{o}}pf},
  \bibinfo{person}{Edward Yang}, \bibinfo{person}{Zachary DeVito},
  \bibinfo{person}{Martin Raison}, \bibinfo{person}{Alykhan Tejani},
  \bibinfo{person}{Sasank Chilamkurthy}, \bibinfo{person}{Benoit Steiner},
  \bibinfo{person}{Lu Fang}, \bibinfo{person}{Junjie Bai}, {and}
  \bibinfo{person}{Soumith Chintala}.} \bibinfo{year}{2019}\natexlab{}.
\newblock \showarticletitle{PyTorch: An Imperative Style, High-Performance Deep
  Learning Library}. In \bibinfo{booktitle}{\emph{Proc. Conference on Neural
  Information Processing Systems, NeurIPS 2019}}. \bibinfo{pages}{8024--8035}.
\newblock


\bibitem[\protect\citeauthoryear{Rahman, Fookes, Baktashmotlagh, and
  Sridharan}{Rahman et~al\mbox{.}}{2019}]%
        {DBLP:conf/wacv/RahmanFBS19}
\bibfield{author}{\bibinfo{person}{Mohammad~Mahfujur Rahman},
  \bibinfo{person}{Clinton Fookes}, \bibinfo{person}{Mahsa Baktashmotlagh},
  {and} \bibinfo{person}{Sridha Sridharan}.} \bibinfo{year}{2019}\natexlab{}.
\newblock \showarticletitle{Multi-Component Image Translation for Deep Domain
  Generalization}. In \bibinfo{booktitle}{\emph{Proc. Winter Conference on
  Applications of Computer Vision, {WACV} 2019}}. \bibinfo{pages}{579--588}.
\newblock


\bibitem[\protect\citeauthoryear{Rahman, Fookes, Baktashmotlagh, and
  Sridharan}{Rahman et~al\mbox{.}}{2020a}]%
        {DBLP:journals/pr/RahmanFBS20}
\bibfield{author}{\bibinfo{person}{Mohammad~Mahfujur Rahman},
  \bibinfo{person}{Clinton Fookes}, \bibinfo{person}{Mahsa Baktashmotlagh},
  {and} \bibinfo{person}{Sridha Sridharan}.} \bibinfo{year}{2020}\natexlab{a}.
\newblock \showarticletitle{Correlation-aware adversarial domain adaptation and
  generalization}.
\newblock \bibinfo{journal}{\emph{Pattern Recognition}}  \bibinfo{volume}{100}
  (\bibinfo{year}{2020}), \bibinfo{pages}{107124}.
\newblock


\bibitem[\protect\citeauthoryear{Rahman, Fookes, Baktashmotlagh, and
  Sridharan}{Rahman et~al\mbox{.}}{2020b}]%
        {DBLP:books/sp/20/RahmanFBS20}
\bibfield{author}{\bibinfo{person}{Mohammad~Mahfujur Rahman},
  \bibinfo{person}{Clinton Fookes}, \bibinfo{person}{Mahsa Baktashmotlagh},
  {and} \bibinfo{person}{Sridha Sridharan}.} \bibinfo{year}{2020}\natexlab{b}.
\newblock \showarticletitle{On Minimum Discrepancy Estimation for Deep Domain
  Adaptation}.
\newblock In \bibinfo{booktitle}{\emph{Domain Adaptation for Visual
  Understanding}}. \bibinfo{pages}{81--94}.
\newblock


\bibitem[\protect\citeauthoryear{Rahmani and Bennamoun}{Rahmani and
  Bennamoun}{2017}]%
        {DBLP:conf/iccv/RahmaniB17}
\bibfield{author}{\bibinfo{person}{Hossein Rahmani} {and}
  \bibinfo{person}{Mohammed Bennamoun}.} \bibinfo{year}{2017}\natexlab{}.
\newblock \showarticletitle{Learning Action Recognition Model from Depth and
  Skeleton Videos}. In \bibinfo{booktitle}{\emph{Proc. International Conference
  on Computer Vision, {ICCV} 2017}}. \bibinfo{pages}{5833--5842}.
\newblock


\bibitem[\protect\citeauthoryear{Saito, Watanabe, Ushiku, and Harada}{Saito
  et~al\mbox{.}}{2018a}]%
        {DBLP:conf/cvpr/SaitoWUH18}
\bibfield{author}{\bibinfo{person}{Kuniaki Saito}, \bibinfo{person}{Kohei
  Watanabe}, \bibinfo{person}{Yoshitaka Ushiku}, {and} \bibinfo{person}{Tatsuya
  Harada}.} \bibinfo{year}{2018}\natexlab{a}.
\newblock \showarticletitle{Maximum Classifier Discrepancy for Unsupervised
  Domain Adaptation}. In \bibinfo{booktitle}{\emph{Proc. {IEEE} Conference on
  Computer Vision and Pattern Recognition, {CVPR} 2018}}.
  \bibinfo{pages}{3723--3732}.
\newblock


\bibitem[\protect\citeauthoryear{Saito, Yamamoto, Ushiku, and Harada}{Saito
  et~al\mbox{.}}{2018b}]%
        {DBLP:conf/eccv/SaitoYUH18}
\bibfield{author}{\bibinfo{person}{Kuniaki Saito}, \bibinfo{person}{Shohei
  Yamamoto}, \bibinfo{person}{Yoshitaka Ushiku}, {and} \bibinfo{person}{Tatsuya
  Harada}.} \bibinfo{year}{2018}\natexlab{b}.
\newblock \showarticletitle{Open Set Domain Adaptation by Backpropagation}. In
  \bibinfo{booktitle}{\emph{Proc. European Conference on Computer Vision,
  {ECCV} 2018}}. \bibinfo{pages}{156--171}.
\newblock


\bibitem[\protect\citeauthoryear{Simonyan and Zisserman}{Simonyan and
  Zisserman}{2014}]%
        {DBLP:conf/nips/SimonyanZ14}
\bibfield{author}{\bibinfo{person}{Karen Simonyan} {and}
  \bibinfo{person}{Andrew Zisserman}.} \bibinfo{year}{2014}\natexlab{}.
\newblock \showarticletitle{Two-Stream Convolutional Networks for Action
  Recognition in Videos}. In \bibinfo{booktitle}{\emph{Proc. Conference on
  Neural Information Processing Systems, NeurIPS 2014}}.
  \bibinfo{pages}{568--576}.
\newblock


\bibitem[\protect\citeauthoryear{Soomro, Zamir, and Shah}{Soomro
  et~al\mbox{.}}{2012}]%
        {DBLP:journals/corr/abs-1212-0402}
\bibfield{author}{\bibinfo{person}{Khurram Soomro},
  \bibinfo{person}{Amir~Roshan Zamir}, {and} \bibinfo{person}{Mubarak Shah}.}
  \bibinfo{year}{2012}\natexlab{}.
\newblock \showarticletitle{{UCF101:} {A} Dataset of 101 Human Actions Classes
  From Videos in The Wild}.
\newblock \bibinfo{journal}{\emph{CoRR}}  \bibinfo{volume}{abs/1212.0402}
  (\bibinfo{year}{2012}).
\newblock


\bibitem[\protect\citeauthoryear{Tran, Bourdev, Fergus, Torresani, and
  Paluri}{Tran et~al\mbox{.}}{2015}]%
        {DBLP:conf/iccv/TranBFTP15}
\bibfield{author}{\bibinfo{person}{Du Tran}, \bibinfo{person}{Lubomir~D.
  Bourdev}, \bibinfo{person}{Rob Fergus}, \bibinfo{person}{Lorenzo Torresani},
  {and} \bibinfo{person}{Manohar Paluri}.} \bibinfo{year}{2015}\natexlab{}.
\newblock \showarticletitle{Learning Spatiotemporal Features with 3D
  Convolutional Networks}. In \bibinfo{booktitle}{\emph{Proc. International
  Conference on Computer Vision, {ICCV} 2015}}. \bibinfo{pages}{4489--4497}.
\newblock


\bibitem[\protect\citeauthoryear{van~der Maaten and Hinton}{van~der Maaten and
  Hinton}{2008}]%
        {JMLR:v9:vandermaaten08a}
\bibfield{author}{\bibinfo{person}{Laurens van~der Maaten} {and}
  \bibinfo{person}{Geoffrey Hinton}.} \bibinfo{year}{2008}\natexlab{}.
\newblock \showarticletitle{Visualizing Data using t-SNE}.
\newblock \bibinfo{journal}{\emph{Journal of Machine Learning Research}}
  \bibinfo{volume}{9}, \bibinfo{number}{86} (\bibinfo{year}{2008}),
  \bibinfo{pages}{2579--2605}.
\newblock


\bibitem[\protect\citeauthoryear{Wang, Wang, Huang, Wang, and Tan}{Wang
  et~al\mbox{.}}{2018}]%
        {DBLP:conf/mm/WangWHWT18}
\bibfield{author}{\bibinfo{person}{Junbo Wang}, \bibinfo{person}{Wei Wang},
  \bibinfo{person}{Yan Huang}, \bibinfo{person}{Liang Wang}, {and}
  \bibinfo{person}{Tieniu Tan}.} \bibinfo{year}{2018}\natexlab{}.
\newblock \showarticletitle{Hierarchical Memory Modelling for Video
  Captioning}. In \bibinfo{booktitle}{\emph{Proc. International Conference on
  Multimedia, {MM} 2018}}. \bibinfo{pages}{63--71}.
\newblock


\bibitem[\protect\citeauthoryear{Wang, Xiong, Wang, Qiao, Lin, Tang, and
  Van~Gool}{Wang et~al\mbox{.}}{2016}]%
        {wang2016temporal}
\bibfield{author}{\bibinfo{person}{Limin Wang}, \bibinfo{person}{Yuanjun
  Xiong}, \bibinfo{person}{Zhe Wang}, \bibinfo{person}{Yu Qiao},
  \bibinfo{person}{Dahua Lin}, \bibinfo{person}{Xiaoou Tang}, {and}
  \bibinfo{person}{Luc Van~Gool}.} \bibinfo{year}{2016}\natexlab{}.
\newblock \showarticletitle{Temporal segment networks: Towards good practices
  for deep action recognition}. In \bibinfo{booktitle}{\emph{Proc. European
  Conference on Computer Vision, {ECCV} 2016}}. \bibinfo{pages}{20--36}.
\newblock


\bibitem[\protect\citeauthoryear{Wang, Song, Wang, Xu, Hu, and Chai}{Wang
  et~al\mbox{.}}{2021b}]%
        {DBLP:conf/icassp/WangSW0HC21}
\bibfield{author}{\bibinfo{person}{Yatian Wang}, \bibinfo{person}{Xiaolin
  Song}, \bibinfo{person}{Yezhen Wang}, \bibinfo{person}{Pengfei Xu},
  \bibinfo{person}{Runbo Hu}, {and} \bibinfo{person}{Hua Chai}.}
  \bibinfo{year}{2021}\natexlab{b}.
\newblock \showarticletitle{Dual Metric Discriminator for Open Set Video Domain
  Adaptation}. In \bibinfo{booktitle}{\emph{Proc. International Conference on
  Acoustics, Speech and Signal Processing, {ICASSP} 2021}}.
  \bibinfo{pages}{8198--8202}.
\newblock


\bibitem[\protect\citeauthoryear{Wang, Luo, Huang, and Baktashmotlagh}{Wang
  et~al\mbox{.}}{2020}]%
        {DBLP:conf/mm/WangLHB20}
\bibfield{author}{\bibinfo{person}{Zijian Wang}, \bibinfo{person}{Yadan Luo},
  \bibinfo{person}{Zi Huang}, {and} \bibinfo{person}{Mahsa Baktashmotlagh}.}
  \bibinfo{year}{2020}\natexlab{}.
\newblock \showarticletitle{Prototype-Matching Graph Network for Heterogeneous
  Domain Adaptation}. In \bibinfo{booktitle}{\emph{Proc. International
  Conference on Multimedia, {MM} 2020}}. \bibinfo{pages}{2104--2112}.
\newblock


\bibitem[\protect\citeauthoryear{Wang, Luo, Qiu, Huang, and
  Baktashmotlagh}{Wang et~al\mbox{.}}{2021a}]%
        {DBLP:journals/corr/abs-2108-11726}
\bibfield{author}{\bibinfo{person}{Zijian Wang}, \bibinfo{person}{Yadan Luo},
  \bibinfo{person}{Ruihong Qiu}, \bibinfo{person}{Zi Huang}, {and}
  \bibinfo{person}{Mahsa Baktashmotlagh}.} \bibinfo{year}{2021}\natexlab{a}.
\newblock \showarticletitle{Learning to Diversify for Single Domain
  Generalization}.
\newblock \bibinfo{journal}{\emph{CoRR}}  \bibinfo{volume}{abs/2108.11726}
  (\bibinfo{year}{2021}).
\newblock


\bibitem[\protect\citeauthoryear{Wei, Cheng, Yu, Zhao, Zhu, and Nie}{Wei
  et~al\mbox{.}}{2019a}]%
        {DBLP:conf/mm/WeiCYZZN19}
\bibfield{author}{\bibinfo{person}{Yinwei Wei}, \bibinfo{person}{Zhiyong
  Cheng}, \bibinfo{person}{Xuzheng Yu}, \bibinfo{person}{Zhou Zhao},
  \bibinfo{person}{Lei Zhu}, {and} \bibinfo{person}{Liqiang Nie}.}
  \bibinfo{year}{2019}\natexlab{a}.
\newblock \showarticletitle{Personalized Hashtag Recommendation for
  Micro-videos}. In \bibinfo{booktitle}{\emph{Proc. International Conference on
  Multimedia, {MM} 2019}}. \bibinfo{pages}{1446--1454}.
\newblock


\bibitem[\protect\citeauthoryear{Wei, Wang, Nie, He, Hong, and Chua}{Wei
  et~al\mbox{.}}{2019b}]%
        {DBLP:conf/mm/WeiWN0HC19}
\bibfield{author}{\bibinfo{person}{Yinwei Wei}, \bibinfo{person}{Xiang Wang},
  \bibinfo{person}{Liqiang Nie}, \bibinfo{person}{Xiangnan He},
  \bibinfo{person}{Richang Hong}, {and} \bibinfo{person}{Tat{-}Seng Chua}.}
  \bibinfo{year}{2019}\natexlab{b}.
\newblock \showarticletitle{{MMGCN:} Multi-modal Graph Convolution Network for
  Personalized Recommendation of Micro-video}. In
  \bibinfo{booktitle}{\emph{Proc. International Conference on Multimedia, {MM}
  2019}}. \bibinfo{pages}{1437--1445}.
\newblock


\bibitem[\protect\citeauthoryear{Yan, Wang, Li, and Qiao}{Yan
  et~al\mbox{.}}{2019}]%
        {DBLP:conf/cvpr/YanWLQ19}
\bibfield{author}{\bibinfo{person}{An Yan}, \bibinfo{person}{Yali Wang},
  \bibinfo{person}{Zhifeng Li}, {and} \bibinfo{person}{Yu Qiao}.}
  \bibinfo{year}{2019}\natexlab{}.
\newblock \showarticletitle{{PA3D:} Pose-Action 3D Machine for Video
  Recognition}. In \bibinfo{booktitle}{\emph{Proc. Conference on Computer
  Vision and Pattern Recognition, {CVPR} 2019}}. \bibinfo{pages}{7922--7931}.
\newblock


\bibitem[\protect\citeauthoryear{Yan, Ding, Li, Wang, Xu, and Zuo}{Yan
  et~al\mbox{.}}{2017}]%
        {DBLP:conf/cvpr/YanDLWXZ17}
\bibfield{author}{\bibinfo{person}{Hongliang Yan}, \bibinfo{person}{Yukang
  Ding}, \bibinfo{person}{Peihua Li}, \bibinfo{person}{Qilong Wang},
  \bibinfo{person}{Yong Xu}, {and} \bibinfo{person}{Wangmeng Zuo}.}
  \bibinfo{year}{2017}\natexlab{}.
\newblock \showarticletitle{Mind the Class Weight Bias: Weighted Maximum Mean
  Discrepancy for Unsupervised Domain Adaptation}. In
  \bibinfo{booktitle}{\emph{Proc. Conference on Computer Vision and Pattern
  Recognition, {CVPR} 2017}}. \bibinfo{pages}{945--954}.
\newblock


\bibitem[\protect\citeauthoryear{Yang, Han, and Wang}{Yang
  et~al\mbox{.}}{2017}]%
        {DBLP:conf/mm/YangHW17}
\bibfield{author}{\bibinfo{person}{Ziwei Yang}, \bibinfo{person}{Yahong Han},
  {and} \bibinfo{person}{Zheng Wang}.} \bibinfo{year}{2017}\natexlab{}.
\newblock \showarticletitle{Catching the Temporal Regions-of-Interest for Video
  Captioning}. In \bibinfo{booktitle}{\emph{Proc. International Conference on
  Multimedia, {MM} 2017}}. \bibinfo{pages}{146--153}.
\newblock


\bibitem[\protect\citeauthoryear{Zhang and Patel}{Zhang and Patel}{2017}]%
        {DBLP:journals/pami/ZhangP17}
\bibfield{author}{\bibinfo{person}{He Zhang} {and} \bibinfo{person}{Vishal~M.
  Patel}.} \bibinfo{year}{2017}\natexlab{}.
\newblock \showarticletitle{Sparse Representation-Based Open Set Recognition}.
\newblock \bibinfo{journal}{\emph{IEEE Transactions on Pattern Analysis and
  Machine Intelligence}} \bibinfo{volume}{39}, \bibinfo{number}{8}
  (\bibinfo{year}{2017}), \bibinfo{pages}{1690--1696}.
\newblock


\bibitem[\protect\citeauthoryear{Zhao, des Combes, Zhang, and Gordon}{Zhao
  et~al\mbox{.}}{2019}]%
        {DBLP:conf/icml/0002CZG19}
\bibfield{author}{\bibinfo{person}{Han Zhao}, \bibinfo{person}{Remi~Tachet des
  Combes}, \bibinfo{person}{Kun Zhang}, {and} \bibinfo{person}{Geoffrey~J.
  Gordon}.} \bibinfo{year}{2019}\natexlab{}.
\newblock \showarticletitle{On Learning Invariant Representations for Domain
  Adaptation}. In \bibinfo{booktitle}{\emph{Proc. International Conference on
  Machine Learning, {ICML} 2019}}. \bibinfo{pages}{7523--7532}.
\newblock


\bibitem[\protect\citeauthoryear{Zhao, Yue, Zhang, Li, Zhao, Wu, Krishna,
  Gonzalez, Sangiovanni-Vincentelli, Seshia, and Keutzer}{Zhao
  et~al\mbox{.}}{2020}]%
        {9238468}
\bibfield{author}{\bibinfo{person}{Sicheng Zhao}, \bibinfo{person}{Xiangyu
  Yue}, \bibinfo{person}{Shanghang Zhang}, \bibinfo{person}{Bo Li},
  \bibinfo{person}{Han Zhao}, \bibinfo{person}{Bichen Wu},
  \bibinfo{person}{Ravi Krishna}, \bibinfo{person}{Joseph~E. Gonzalez},
  \bibinfo{person}{Alberto~L. Sangiovanni-Vincentelli},
  \bibinfo{person}{Sanjit~A. Seshia}, {and} \bibinfo{person}{Kurt Keutzer}.}
  \bibinfo{year}{2020}\natexlab{}.
\newblock \showarticletitle{A Review of Single-Source Deep Unsupervised Visual
  Domain Adaptation}.
\newblock \bibinfo{journal}{\emph{IEEE Transactions on Neural Networks and
  Learning Systems}} (\bibinfo{year}{2020}).
\newblock


\bibitem[\protect\citeauthoryear{Zhou, Andonian, Oliva, and Torralba}{Zhou
  et~al\mbox{.}}{2018}]%
        {DBLP:conf/eccv/ZhouAOT18}
\bibfield{author}{\bibinfo{person}{Bolei Zhou}, \bibinfo{person}{Alex
  Andonian}, \bibinfo{person}{Aude Oliva}, {and} \bibinfo{person}{Antonio
  Torralba}.} \bibinfo{year}{2018}\natexlab{}.
\newblock \showarticletitle{Temporal Relational Reasoning in Videos}. In
  \bibinfo{booktitle}{\emph{Proc. European Conference on Computer Vision,
  {ECCV} 2018}}. \bibinfo{pages}{831--846}.
\newblock


\end{thebibliography}

\appendix

\end{document}